%% file: main_arxiv_v2_extension_version.tex
\documentclass[letterpaper]{article} 
\usepackage{aaai23}

\usepackage{times}  
\usepackage{helvet}  
\usepackage{courier}  
\usepackage[hyphens]{url}  
\usepackage{graphicx} 
\usepackage{natbib}  
\usepackage{caption} 
\title{Self-Supervised Primal-Dual Learning for Constrained Optimization}
\author {
    Seonho~Park, Pascal~Van~Hentenryck
}
\affiliations {
    H. Milton Stewart School of Industrial \& Systems Engineering, Georgia Institute of Technology\\
    seonho.park@gatech.edu, pascal.vanhentenryck@isye.gatech.edu
}

\input{commands}
\usepackage{hyperref}
\hypersetup{pdfauthor   = {Seonho Park, Pascal Van Hentenryck},
            pdftitle    = {Self-Supervised Primal-Dual Learning for Constrained Optimization},
            pdfsubject  = {Arxiv v1},
            pdfkeywords = {Primal-Dual Learning, Constrained Optimization Learning, Augmented Lagrangian Method, Quadratic Program, Quadratically Constrained Quadratic Program, AC-OPF, Optimal Power Flow},
            colorlinks=true,
            linkcolor=red,
            filecolor=blue,      
            urlcolor=blue,
            citecolor=blue
            }

\nocopyright

\begin{document}

\maketitle

\begin{abstract}
    This paper studies how to train machine-learning models that
    directly approximate the optimal solutions of constrained
    optimization problems. This is an
    empirical risk minimization under constraints, which is
    challenging as training must balance optimality and
    feasibility conditions. Supervised learning methods often approach
    this challenge by training the model on a large collection of
    pre-solved instances. This paper takes a different route and
    proposes the idea of Primal-Dual Learning (PDL), a self-supervised
    training method that does not require a set of pre-solved
    instances or an optimization solver for training and
    inference. Instead, PDL mimics the trajectory of an Augmented
    Lagrangian Method (ALM) and jointly trains primal and dual neural
    networks. Being a primal-dual method, PDL uses instance-specific
    penalties of the constraint terms in the loss function used to
    train the primal network. Experiments show that, on a set of
    nonlinear optimization benchmarks, PDL typically exhibits
    negligible constraint violations and minor optimality gaps, and is
    remarkably close to the ALM optimization. PDL also demonstrated
    improved or similar performance in terms of the optimality gaps,
    constraint violations, and training times compared to
    existing approaches.
\end{abstract}

\section{Introduction}\label{sec:intro}

\noindent
This paper considers constrained optimization problems which are
ubiquitous in many disciplines including power systems, supply chains,
transportation and logistics, manufacturing, and design. Some of these
problems need to be solved within time limits to meet business
constraints. This is the case for example of market-clearing
optimizations in power systems, sensitivity analyses in manufacturing,
and real-time transportation systems to name only a few. In many
cases, these optimization problems must be solved repeatedly for
problem instances that are closely related to each other (the so-called
structure hypothesis \cite{bengio2021machine}). These observations have stimulated
significant interest in recent years at the intersection of machine
learning and optimization.

This paper is concerned with one specific approach in this rich
landscape: the idea of learning the input/output mapping of an
optimization problem. This approach has raised significant interest in
recent years, especially in power systems. Supervised machine learning
has been the most prominent approach to tackling these problems (e.g.,
\cite{zamzam2020learning,fioretto2020predicting,chatzos2021spatial,kotary2021learning,chatzos2020high,kotary2022fast,PSCC2022JITL}). Moreover,
to balance feasibility and optimality, existing methods may perform a
Lagrangian relaxation of the constraints.  Supervised learning relies
on a collection of pre-solved instances usually obtained through
historical data and data generation.

These supervised methods have achieved promising results on a variety
of problems. They also have some limitations including the need to
generate the instance data which is often a time-consuming and
resource-intensive process. Moreover, the data generation is often
subtle as optimization problems may admit multiple solutions and
exhibit many symmetries. As a result, the data generation must be
orchestrated carefully to simplify the learning process
\cite{kotary2021learning}. Recently, \citet{donti2021dc3} have
proposed a self-supervised learning method, called DC3, that addresses
these limitations: it bypasses the data generation process by training
the machine-learning model with a loss function that captures the
objective function and constraints simultaneously.

Existing approaches are also primal methods: they learn a network to
predict the values of the decision variables. As a result, the
constraint multipliers they use to balance feasibility and optimality
are not instance-specific: they are aggregated for each constraint
over all the instances. This aggregation limits the capability of the
learning method to ensure feasibility. For these reasons, dedicated
procedures have been proposed to restore feasibility in a
post-processing step (e.g.,
\cite{zamzam2020learning,velloso2021combining,chen2021enforcing,fioretto2020predicting}).
This repair process however may lead to sub-optimal solutions.

This paper is an attempt to address these limitations for some classes
of applications. It proposes Primal-Dual Learning (PDL), a self-supervised primal-dual
learning method to approximate the input/output mapping of a
constrained optimization problem. Being a self-supervised method, PDL
does not rely on a set of pre-solved instances. Moreover, being a
primal-dual method, PDL is capable of using instance-specific
multipliers for penalizing constraints. The key idea underlying PDL is
to mimic the trajectories of an Augmented Lagrangian Method (ALM) and
to jointly train primal and dual neural networks. As a result, during
each iteration, the training process learns the primal and dual
iteration points of the ALM algorithm. Eventually, these iteration
points, and hence the primal and dual networks, are expected to
converge to the primal and dual solutions of the optimization problem.

The effectiveness of PDL is demonstrated on a number of benchmarks
including convex quadratic programming (QP) problems and their
non-convex QP variants, quadratic constrained
quadratic programming (QCQP), and optimal power flows (OPF) for energy
systems. The results highlight that PDL finds solutions with small
optimality gaps and negligible constraint violations. Its
approximations are remarkably close to those of optimization with ALM
and, on the considered test cases, PDL exhibits improved accuracy
compared to supervised methods and primal self-supervised approaches.

In summary, the key contributions of PDL are summarized as follows:
\begin{itemize}
\item PDL proposes a self-supervised learning method for learning
the input/output mapping of constrained optimization problems. It does
so by mimicking the trajectories of the ALM without requiring pre-solved 
instances or the use of an optimization solver.

\item PDL is a primal-dual method: it jointly trains two networks to
approximate the primal and dual solutions of the underlying optimization.
PDL can leverage the dual network to impose instance-specific 
multipliers/penalties for the constraints in the loss function.

\item Experimental results show that, at inference time, PDL obtains
solutions with negligible constraint violations and minor optimality gaps
on a collection of benchmarks that include QP, QCQP, and OPF test cases.
\end{itemize}

\noindent
The rest of this paper is organized as follows. Section
\ref{sec:related} presents the related work at a high level. Section
\ref{sec:prelim} presents the paper notation and learning goal,
and it reviews existing approaches in more detail. Section
\ref{sec:method} presents PDL and Section \ref{sec:experiment} reports
the experimental results. Section \ref{sec:conclusion} concludes the
paper.

\section{Related Work}
\label{sec:related}

This section gives a broad overview of related work. Detailed
presentations of the key related work are introduced in
Section \ref{sec:prelim}. The use of machine learning for optimization
problems can be categorized into two main threads (See the survey by 
\citet{bengio2021machine} for a broad overview of this area): \romannum{1}) {\em
Learning to optimize} where machine learning is used to improve an
optimization solver by providing better heuristics or branching rules
for instance \cite{chen2021learning,liu2022learning}; \romannum{2})
{\em optimization proxies/surrogates} where the machine-learning model
directly approximates the input/output mapping of the optimization
problem \cite{vesselinova2020learning,kotary2021end}.
    
Reinforcement learning has been mostly used for the first category
including learning how to branch and how to apply cutting planes in
mixed integer programming
(e.g., \cite{khalil2016learning,tang2020reinforcement,liu2022learning}, and how to
derive heuristics for combinatorial optimization problems on graphs
(e.g., \cite{khalil2017learning}).

Supervised learning has been mostly used for the second category.  For
example, \citet{fioretto2020predicting}, \citet{chatzos2020high},
and \citet{zamzam2020learning} propose deep-learning models for
optimal power flow that approximate generator setpoints. These
supervised approaches require a set of pre-solved instances obtained
from historical data and complemented by a data augmentation process
that uses an optimization solver. This data augmentation process can
be very time-consuming and resource-intensive, and may not be
practical for certain type of applications (e.g., \cite{PSCC2022JITL}).

\citet{donti2021dc3} propose a self-supervised learning approach that
avoids the need for pre-solved instances, using the objective function
and constraint violations to train the neural network directly.

\section{Preliminaries}
\label{sec:prelim}
\subsection{Notations}
\label{ssec:notations} 

Bold lowercase notations are used for vectors or sets of functions,
and bold uppercase notations are used for matrices.
A capital calligraphic notation such
as $\cG$ represents a set, a distribution, or a loss
function. $\norm{\cdot}$ represents an arbitrary norm.
    
\subsection{Assumptions and Goal}
\label{ssec:assumption_goal}

This paper considers a constrained optimization problem
$P_{\bx}(\by)$ of the form:
\begin{equation}
\label{eq:formulation}
\begin{aligned}
&P_\bx(\by):&&\minimize_{\by\in\bY} && f_\bx(\by),\\
&&&\text{\textbf{subject to }} && g_{\bx,j}(\by) \leq 0,\;\;\forall j\in\cG,\\ 
&&&                            && h_{\bx,j}(\by) = 0,\;\; \forall j\in\cH.
\end{aligned}
\end{equation}

\noindent
$\bx\in\RR^p$ represents instance parameters that determine the
functions $f_\bx$, $\bg_\bx$, and $\bh_\bx$.
Each $\bx$ thus defines an instance of the optimization problem
\eqref{eq:formulation}. $\cG$ and $\cH$ are the indices sets of the
inequality and equality functions, respectively.  $\bY$ is the
variable domain, i.e., a closed subset of $\RR^n$ defined by some
bound constraints.  The functions $f_\bx$, $\bg_\bx$, and $\bh_\bx$ are smooth
over $\bY$, and possibly nonlinear and non-convex.

The goal is to approximate the mapping from $\bx$ to an
optimal solution $\ygt$ of the problem $P_{\bx}(\by)$
(Eq.~\eqref{eq:formulation}). This work restricts the potential
approximate mappings to a family of neural nets $N_\theta$
parameterized by trainable parameters $\theta$.  The model $N_\theta$
is trained offline and, at inference time, it is expected to yield, in
real-time, a high-quality approximation to an optimal solution.  This
paper measures the quality of an approximation using optimality gaps
and maximum constraint violations over unseen test instances.

\subsection{Supervised Learning}
\label{ssec:supervised_learning}

This subsection revisits some supervised-learning schemes to meet the
above requirements. These methods need the ground truth, i.e., an actual
optimal solution for each instance parameters $\bx$ that can be obtained
by solving problem $P_{\bx}(\by)$ with an optimization solver. Given
the dataset consisting of pairs $\{\bx,\ygt\}$, the neural net
$N_\theta$ strives to yield an output $\by$ approximating the ground
truth $\ygt$ as accurately as possible.\footnote{The loss functions in
  what follows generalize naturally to mini-batches of instances.}
        
\paragraph{The \Naive{} Approach}

The \naive{} supervised-learning approach (e.g.,
\cite{zamzam2020learning}) minimizes a simple and intuitive loss
function such as
\begin{equation}\label{eq:naive_supervised_loss}
\cL_{\text{\naive{}}} = \norm{\by -\ygt}.
\end{equation}
\noindent
MAE and MSE have been widely used as loss functions. The main
limitation of the \naive{} approach is that it may provide approximations
that violate constraints significantly.

\paragraph{The Supervised Penalty Approach}

The supervised penalty approach uses a loss function that includes
terms penalizing constraint violations (e.g., \cite{nellikkath2022physics})
\begin{equation}\label{eq:penalty_supervised_loss} 
\begin{aligned} 
&\cL_{\text{\naive{}}}
+ \sum_{j\in\cG}{\rho_{g,j} \viol{g_{\bx,j}(\by)}}
+ \sum_{j\in\cH}{\rho_{h,j}  \viol{h_{\bx,j}(\by)}}
\end{aligned}
\end{equation}
where $g_{\bx,j}$ and $h_{\bx,j}$ are the $j^{th}$ inequality and
equality constraints, respectively, and $\viol{\cdot}$ is any
violation penalty function.  Each constraint also has a penalty
coefficient $\rho > 0$.
        
\paragraph{The Lagrangian Duality Method}

It is not an obvious task to choose penalty coefficients (or
Lagrangian multipliers) to minimize constraint violations and balance
them with the objective function. \citet{fioretto2020predicting}
propose a Lagrangian Duality (LD) approach to determine suitable
Lagrangian multipliers. They employ a subgradient method that updates the
multipliers every few epochs using the rules:
\begin{equation*}\label{eq:ld_penalty_update}
\begin{aligned}
\rho_{g,j} & \leftarrow \rho_{g,j}+\gamma \viol{g_{\bx,j}(\by)},\;\forall j\in\cG, \;\text{and} \\
\rho_{h,j} &\leftarrow \rho_{h,j}+\gamma \viol{h_{\bx,j}(\by)},\; \forall j\in\cH, 
\end{aligned} 
\end{equation*}
\noindent
where the hyperparameter $\gamma>0$ represents the step size for updating
$\rho$. 

\subsection{Self-Supervised Learning}
\label{ssec:self_supervised_learning}

Obtaining $\ygt$ for a large collection of instances is a
time-consuming and resource-intensive task. In addition, the data
augmentation process should proceed carefully since modern solvers are
often randomized and may yield rather different optimal solutions to
similar optimization problems
\cite{kotary2021learning}. Self-supervised learning approaches remedy
this limitation by learning the neural network directly without using
any pre-solved instances. Instead, they use a loss function that
includes the objective function of the optimization problem.

\paragraph{The Self-Supervised Penalty Approach}

The self-supervised penalty approach uses the loss function
\begin{equation}
\label{eq:penalty_selfsupervised_loss}
\begin{aligned}
f_\bx(\by) + \sum_{j\in\cG}{\rho_{g,j}\viol{g_{\bx,j}(\by)}} + \sum_{j\in\cH}{\rho_{h,j}\viol{h_{\bx,j}(\by)}}.
\end{aligned}
\end{equation}
This loss function resembles the supervised penalty approach, except
that the objective function term replaces the norm term that measures 
the proximity to the ground truth.

\paragraph{DC3}

DC3 (Deep Constraint Completion and Correction) is a self-supervised
learning method proposed by \citet{donti2021dc3}. It uses the same
loss as the self-supervised penalty approach
(Eq.~\eqref{eq:penalty_selfsupervised_loss}), but also employs completion
and correction steps to ensure feasibility. In particular, the core
neural network only approximates a subset of the variables. It is then
followed by a completion step that uses implicit differentiation to 
compute a variable assignment that satisfies the
equality constraints. The correction step then uses gradient descent
to correct the output and satisfy the inequality constraints. Those
steps ensure that a feasible solution is found. However, the obtained solution
may be sub-optimal, as shown in Section \ref{sec:experiment}.

\section{Self-Supervised Primal-Dual Learning}
\label{sec:method}

This section is the core of the paper. It presents a self-supervised
Primal-Dual Learning (PDL) that aims at combining the benefits of
self-supervised and Lagrangian dual methods. The key characteristics
of PDL can be summarized as follows.

\begin{itemize}
\item \textbf{Self-Supervised Training Process Inspired by ALM} The training process of PDL mimics an Augmented Lagrangian Method (ALM) \cite{hestenes1969multiplier,powell1969method,rockafellar1974augmented,bertsekas2014constrained}.
\item \textbf{Primal and Dual Approximations} PDL jointly trains two independent networks: one for approximating primal variables and another for approximating dual variables that can then serve as Lagrangian multipliers in the primal training. 
\item \textbf{Instance-Specific Lagrangian Multipliers} Contrary to the supervised methods, PDL uses instance-specific Lagrangian multipliers, yielding a fine-grained balance between constraint satisfaction and optimality conditions. 
\end{itemize}

\begin{figure}[!t]
\centering
\includegraphics[width=.99\columnwidth]{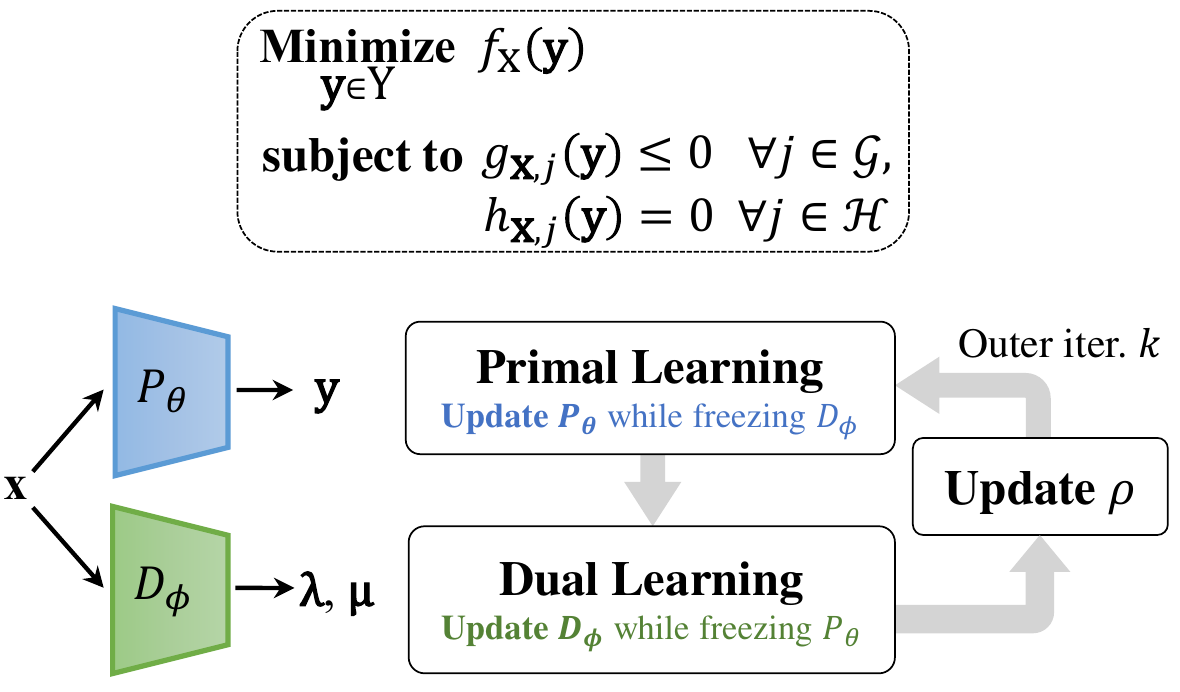}
\caption{Schematic overview of PDL for training both primal and dual neural networks that output the optimal solutions of the constrained optimization problems.}
\label{fig:pdl_overview}
\end{figure}

\noindent
A high-level overview of PDL is presented in
Figure~\ref{fig:pdl_overview}. It highlights that PDL alternates between
primal learning, dual learning, and updating a penalty coefficient at
each outer iteration. Given instances $\bx$, the \emph{primal network}
$P_\theta$ outputs the primal solution estimates $\by$, i.e.,
$P_\theta(\bx) = \by$, whereas the \emph{dual network} $D_\phi$
provides the dual solution estimates $\lamg,\lamh$ for equalities and
inequalities, i.e., $D_\phi(\bx)=\{\lamg, \lamh\}$.  The trainable
parameters $\theta$ and $\phi$, associated with the primal and dual
networks, respectively, are learned during training. Specifically, the
\emph{primal learning} updates $P_\theta$ while $D_\phi$ is
frozen. Conversely, the \emph{dual learning} updates $D_\phi$ while
$P_\theta$ is frozen.  Observe that PDL obtains both primal and dual
estimates at inference time and is able to leverage any neural network
architecture. The rest of this section provides all the details of
PDL.

\subsection{Primal Learning}
\label{ssec:primal_learning}

The loss function for training the primal network is the objective
function of the primal subproblem of the ALM: it combines a Lagrangian relaxation with a penalty on
constraint violations. For each outer iteration $k$, 
the parameters $\theta$ of the primal network $P_\theta$ are trained while freezing $D_\phi$. 
The loss function of the primal training is given by
\begin{equation}\label{eq:pdl_primal_loss}
\begin{aligned}
\cL_{\text{p}}\left( \by \middle| \lamg,\lamh \right) \!&= f_\bx(\by) \!+\! \lamg^T\mathbf{g}_\bx(\by) \!+\! \lamh^T\mathbf{h}_\bx(\by)+\\
            &\frac{\rho}{2}\!\!\left( \sum_{j\in\cG}\viol{g_{\bx,j}(\by)} \!+\! \sum_{j\in\cH}\viol{h_{\bx,j}(\by)} \!\right),
\end{aligned}
\end{equation}
\noindent
where $\lamg$ and $\lamh$ are the output of the frozen dual network
$D_\phi$, i.e., the dual solution estimates for the inequality and
equality constraints, respectively. {\em Note that these dual estimates are
instance-specific: for instance $\bx$, the dual estimates are given by
$D_\phi(\bx)$.} The dual estimates are initialized to zero at the first outer
iteration (which is readily achievable by imposing the initial weights and bias
parameters of the final layer of $D_\phi$ to be zero). 
Observe that the initial dual estimate setting induces the primal loss 
function~\eqref{eq:pdl_primal_loss} is identical to the penalty 
loss~\eqref{eq:penalty_selfsupervised_loss}.
Also, this setting is mostly used in ALM itself.
In Eq.~\eqref{eq:pdl_primal_loss}, following the ALM, the violation
functions for the inequality and equality constraints are defined as
$\viol{g(\by)}=\max\{g(\by),0\}^2$ and $\viol{h(\by)}=h(\by)^2$.
        
\subsection{Dual Learning}
\label{ssec:dual_learning}

The dual learning is particularly interesting. PDL does not store the
dual estimates for all instances for ensuring efficiency and scalability. 
Instead, it trains the dual network $D_\phi$ to estimate them on a need basis.
This training is again inspired by the ALM algorithm. At outer iteration
$k$, the ALM updates the dual estimates using a proximal subgradient step:
\begin{equation}
\label{eq:alm_dual}
\begin{aligned}
&\lamg_{k+1} \leftarrow \max\{\lamg_k+\rho\mathbf{g}(\by),0\},\\
&\lamh_{k+1} \leftarrow \lamh_k+\rho\mathbf{h}(\by).
\end{aligned}
\end{equation}
This update step suggests the targets for updating $D_\phi$. However,
the dual training requires some care since the update rules refer both
to the current estimates of the dual values, and to their new
estimates. For this reason, at outer iteration $k$, the dual learning
first copies the dual network into a ``frozen'' network $D_{\phi_k}$
that will provide the current estimates $\lamg_k$ and $\lamh_k$. 
The dual learning can then update the dual network to yield a
new estimate that is close to the RHS of the ALM dual update rule (Eq.~\eqref{eq:alm_dual}). 
Specifically, the dual network $D_\phi$ is trained by
minimizing the following dual loss function:
\begin{equation}\label{eq:pdl_dual_loss}
\begin{aligned}
    \cL_{\text{d}}\!\left(\!\lamg,\lamh \middle| \by, \lamg_k,\lamh_k \!\right) \!&=\!\norm{\lamg \!-\! \left(\max\{\lamg_k \!+\! \rho \mathbf{g}_\bx(\by),0\}\right)}\\
                                                                                  &+\norm{\lamh \!-\! \left(\!\lamh_k \!+\! \rho \mathbf{h}_\bx(\by)\right)}.
\end{aligned}
\end{equation}
\noindent
where $\lamg_k$ and $\lamh_k$ are the outputs of the frozen network $D_{\phi_k}$ 
and $\by$ is the output of the primal network $P_\theta$. Note that the
primal network $P_\theta$ is also frozen during the dual learning.
Moreover, by minimizing the dual loss function
(Eq.~\eqref{eq:pdl_dual_loss}), it is expected that, for each instance,
the dual network yields the dual solution estimates that are close to
those obtained in the ALM algorithm after applying the first-order proximal rule
(Eq.~\eqref{eq:alm_dual}).

\begin{algorithm}[!t]
\caption{Primal-Dual Learning (PDL)}\label{alg:pdl}
\textbf{Parameter}: 
    Initial penalty coefficient $\rho$, 
    Maximum outer iteration $K$, 
    Maximum inner iteration $L$, 
    Penalty coefficient updating multiplier $\alpha$, 
    Violation tolerance $\tau$, 
    Upper penalty coefficient safeguard $\rhomax$ \\ 
\textbf{Input}: Training dataset $\dataset$\\
\textbf{Output}: learned primal and dual networks $P_\theta$, $D_\phi$
\begin{algorithmic}[1] 
\For{$k \in \{1,\dots,K\}$}
    \For{$l \in \{1,\dots,L\}$} \Comment{Primal Learning}
        \State Update $P_\theta$ using $\nabla_\theta \cL_{\text{p}}$ (See Eq.~\eqref{eq:pdl_primal_loss})
    \EndFor
    \State Calculate $v_k$ as Eq.~\eqref{eq:rho_violations}
    \State Define $D_{\phi_k}$ by copying $D_\phi$
    \For{$l \in \{1,\dots,L\}$} \Comment{Dual Learning}
        \State Update $D_\phi$ using $\nabla_\phi \cL_{\text{d}}$ (See Eq.~\eqref{eq:pdl_dual_loss})
    \EndFor
    \State Update $\rho$ with Eq.~\eqref{eq:update_rho-main}
\EndFor
\State \textbf{return} $\theta$ and $\phi$
\end{algorithmic}
\end{algorithm}

\subsection{Updating the Penalty Coefficient $\rho$}
\label{ssec:update_rho}

When constraint violations are severe, it is desirable to increase the
penalty coefficient $\rho$, which is also used as a step size for
updating the duals, in order to penalize violations more.  PDL adopts
the penalty coefficient updating rule from
\cite{andreani2008augmented} but slightly modifies it to consider all
instances in the training dataset.  At each outer iteration $k$, the
maximum violation $v_k$ is calculated as
\begin{align}
    v_k = \max_{\bx\sim\dataset}\left\{\max\{ \linfnorm{\mathbf{h}_\bx(\by)},\linfnorm{\bm{\sigma_\bx(\by)}} \}\right\},\label{eq:rho_violations} \\
    \text{where }\sigma_{\bx,j}(\by)=\max\{g_{\bx,j}(\by), -\frac{\lambda_{k,j}}{\rho}\} ,\;\forall j\in\cG \nonumber\label{eq:rho_sigma}
\end{align}
\noindent
and $\dataset=\{\bx^{(i)}\}_{i=1}^N$ is the training dataset.
Informally speaking, the vector $\bm{\sigma}$ represents the
infeasibility and complementarity for the inequality
constraints \cite{andreani2008augmented}. At every outer iteration $k>1$,
the penalty coefficient is increased when the maximum violation $v_k$
is greater than a tolerance value $\tau v_{k-1}$ as follows:
\begin{equation}
\label{eq:update_rho-main}
\rho \leftarrow \min\{\alpha\rho,\rhomax\}\;\text{if
        } v_k > \tau v_{k-1},
\end{equation}
\noindent
where $\tau\in(0,1)$ is a tolerance to determine the update of the
penalty coefficient $\rho$, $\alpha>1$ is an update multiplier, and
$\rhomax$ is the upper safeguard of $\rho$.

\subsection{Training}
\label{ssec:training}

The overall training procedure of PDL is detailed in
Algorithm~\ref{alg:pdl}.  The training alternates between the updates
of the primal and dual networks. Unlike conventional training
processes, the PDL training procedure contains outer and inner
iterations.  In an outer iteration, the two inner loops tackle the
primal and dual learning subproblems. Each inner iteration samples a
mini-batch and updates the trainable parameters ($\theta$ or $\phi$).
{\em Note that each outer iteration aims at estimating the primal and
  dual iteration points of the ALM algorithm. Eventually, these
  iteration points are expected to converge to the primal and dual
  solutions on the training instances and also provide the optimal
  solution estimates on unseen testing instances.}

\section{Experiments}
\label{sec:experiment}

This section demonstrates the effectiveness of PDL on a number of
constrained optimization problems. The performance of PDL
is measured by its optimality gaps and maximum constraint
violations, which quantify both optimality and feasibility,
respectively.  PDL is compared to the supervised (SL) and self-supervised (SSL) baselines
introduced in Section~\ref{sec:prelim}, i.e.,
\begin{itemize}
\item \textbf{(Supervised) \naive{} MAE, \naive{} MSE:} they respectively use the l$_1$-norm and the l$_2$-norm between the optimal solution estimates and the actual optimal solutions as loss functions (See Eq.~\eqref{eq:naive_supervised_loss}).

\item \textbf{(Supervised) MAE+Penalty, MSE+Penalty, LD:} on top of \naive{}-MAE (or \naive{}-MSE), these methods add terms to penalize constraint violations (See Eq.~\eqref{eq:penalty_supervised_loss}). For LD, l$_1$-norm is associated as used in \cite{chatzos2020high}.

\item \textbf{(Self-Supervised) Penalty} this method uses a loss function
that represents the average of the penalty functions given in Eq.~\eqref{eq:penalty_selfsupervised_loss} over the training instances.

\item \textbf{(Self-Supervised) DC3} The self-supervised method proposed by \citet{donti2021dc3}.
\end{itemize}

\noindent
Note that DC3 is only tested on QP problems because the publicly
available code\footnote{\url{https://github.com/locuslab/DC3}} is only
applicable to those problems.  Also, the penalty function in
MAE(MSE)+Penalty(SL) and Penalty(SSL) uses absolute violations i.e.,
$\nu(a)=|a|$ for equalities and $\nu(a)=\max\{a,0\}$ for inequalities.

Note that PDL is generic and agnostic about the underlying neural
network architectures.  For comparison purposes with respect to the training methodology, 
the experimental evaluations are based on the simple fully-connected neural networks
followed by ReLU activations.
The implementation is based on PyTorch and the training
was conducted using a Tesla RTX6000 GPU on a machine with Intel Xeon
2.7GHz. For training the models, the Adam optimizer \cite{kingma2014adam} 
with the learning rate of $1\mathrm{e}{\textrm{-}4}$ was used.
Other hyperparameters of PDL and the baselines were tuned using a grid
search.  The hyperparameter settings and architectural design are
detailed in Appendix~\ref{appx:exp_settings}.

\pagebreak 
\subsection{Performance Results}

\subsubsection{Convex QP \& Non-convex QP Variant}

\begin{table*}[!t]
\centering
\small
\begin{tabular}{@{}ll|llllll@{}}
\toprule
Method               & Type & Obj. & Opt. Gap(\%) & Max eq. & Max ineq. & Mean eq. & Mean ineq.\\ \midrule
&&\multicolumn{6}{c}{convex QP (Eq.~\eqref{eq:simple_qp})}   \\\midrule
Optimizer(OSQP) &                    & -15.047        &   -               & 0.000              & 0.001              & 0.000        & 0.000        \\
Optimizer(ALM)  &                    & -15.046        & 0.003             & 0.000              & 0.000              & 0.000        & 0.000        \\ \midrule
PDL(ours)       &\multirow{3}{*}{SSL}& -15.017(0.009) & 0.176(0.054)      & 0.005(0.001)       & 0.001(0.000)       & 0.002(0.000) & 0.000(0.000) \\
Penalty         &                    & -15.149(0.003) & 0.680(0.017)      & \ub{0.048}(0.003)  & \ub{0.030}(0.003)  & 0.013(0.000) & 0.004(0.001) \\
DC3             &                    & -14.112(0.015) & \ub{6.219}(0.098) & 0.000(0.000)       & 0.000(0.000)       & 0.000(0.000) & 0.000(0.000) \\ \midrule
\Naive{} MAE    &\multirow{5}{*}{SL} & -15.051(0.003) & 0.046(0.012)      & \ub{0.025}(0.002)  & \ub{0.018}(0.003)  & 0.008(0.001) & 0.002(0.000) \\
\Naive{} MSE    &                    & -15.047(0.001) & 0.122(0.025)      & \ub{0.018}(0.001)  & \ub{0.023}(0.004)  & 0.006(0.000) & 0.003(0.001) \\
MAE+Penalty     &                    & -15.043(0.004) & 0.093(0.013)      & \ub{0.031}(0.001)  & \ub{0.016}(0.001)  & 0.010(0.000) & 0.002(0.000) \\
MSE+Penalty     &                    & -15.047(0.001) & 0.059(0.004)      & \ub{0.026}(0.003)  & \ub{0.016}(0.001)  & 0.008(0.001) & 0.002(0.000) \\
LD              &                    & -15.043(0.006) & 0.093(0.015)      & \ub{0.033}(0.001)  & \ub{0.016}(0.001)  & 0.011(0.001) & 0.002(0.000) \\ \midrule\midrule
&&\multicolumn{6}{c}{non-convex QP variant (Eq.~\eqref{eq:nonconvex_qp})}\\ \midrule
Optimizer(IPOPT)&                    & -11.592        &   -               & 0.000             & 0.000             & 0.000        & 0.000        \\
Optimizer(ALM)  &                    & -11.592        & 0.002             & 0.000             & 0.000             & 0.000        & 0.000        \\ \midrule
PDL (ours)      &\multirow{3}{*}{SSL}& -11.552(0.006) & 0.324(0.051)      & 0.004(0.001)      & 0.001(0.000)      & 0.001(0.000) & 0.000(0.000) \\
Penalty         &                    & -11.654(0.002) & 0.532(0.015)      & \ub{0.042}(0.002) & \ub{0.027}(0.001) & 0.011(0.000) & 0.003(0.000) \\
DC3             &                    & -11.118(0.017) & \ub{4.103}(0.151) & 0.000(0.000)      & 0.000(0.000)      & 0.000(0.000) & 0.000(0.000) \\ \midrule
\Naive{} MAE    &\multirow{5}{*}{SL} & -11.593(0.004) & 0.044(0.021)      & \ub{0.020}(0.002) & \ub{0.022}(0.006) & 0.006(0.000) & 0.001(0.001) \\
\Naive{} MSE    &                    & -11.593(0.002) & 0.031(0.008)      & \ub{0.017}(0.000) & \ub{0.029}(0.010) & 0.005(0.000) & 0.002(0.001) \\
MAE+Penalty     &                    & -11.591(0.004) & 0.073(0.016)      & \ub{0.033}(0.001) & \ub{0.016}(0.002) & 0.010(0.000) & 0.001(0.000) \\
MSE+Penalty     &                    & -11.593(0.001) & 0.050(0.002)      & \ub{0.023}(0.002) & \ub{0.017}(0.001) & 0.007(0.001) & 0.001(0.000) \\
LD              &                    & -11.593(0.005) & 0.072(0.005)      & \ub{0.032}(0.001) & \ub{0.017}(0.002) & 0.010(0.000) & 0.001(0.000) \\ \bottomrule
\end{tabular}
\caption{Performance results of the self-supervised (SSL) and supervised learning (SL) schemes for the convex QP problem (Top) and non-convex QP problem (Bottom) with $n=100$, $n_{\text{eq}}=n_{\text{ineq}}=50$ on 833 test instances. Std. dev. in parenthesis is evaluated across 5 independent runs. Underlined numbers denote poor results (opt. gap over 1\% or max violation over 0.01).}
\label{tab:result_qp}
\end{table*}
        
This benchmark uses convex QP problems and their non-convex variants
proposed for evaluating DC3 \cite{donti2021dc3}. The experiments
followed the same experimental protocol as in that paper.  Given
various $\bx\in\RR^{n_{\text{eq}}}$, the mathematical formulation for
both cases reads as
\begin{align}
    \min_{\by\in\RR^n} \frac{1}{2}\by^T\mathbf{Q}\by\!+\!\mathbf{r}^T\by, \text{ s.t. } \mathbf{A}\by\!=\!\bx,\;\mathbf{G}\by\leq\mathbf{h},\label{eq:simple_qp}\\
    \min_{\by\in\RR^n} \frac{1}{2}\by^T\mathbf{Q}\by\!+\!\mathbf{r}^T\!\sin (\by), \text{ s.t. } \mathbf{A}\by\!=\!\bx,\,\mathbf{G}\by\leq\mathbf{h},\label{eq:nonconvex_qp}
\end{align}
\noindent
where $\mathbf{Q}\!\in\!\RR^{n\times n}\!\succeq\!0$,
$\mathbf{r}\!\in\!\RR^{n}$, $\mathbf{A}\!\in\!\RR^{n_{\text{eq}}\times
  n}$, $\mathbf{G}\!\in\!\RR^{n_{\text{ineq}}\times n}$, and
$\mathbf{h}\!\in\!\RR^{n_{\text{ineq}}}$. $n$, $n_{\text{eq}}$, and
$n_{\text{ineq}}$ denote the number of variables, equations, and
inequality constraints, respectively.  Eq.~\eqref{eq:nonconvex_qp} is
a non-convex variant of the convex QP (Eq.~\eqref{eq:simple_qp})
obtained by adding the element-wise sine operation.  
Overall, $10,000$ instances were
generated and split into training/testing/validation datasets with the
ratio of $10\!\!:\!\!1\!\!:\!\!1$.
        
\begin{figure}[t!]
    \centering
    \includegraphics[width=.85\columnwidth]{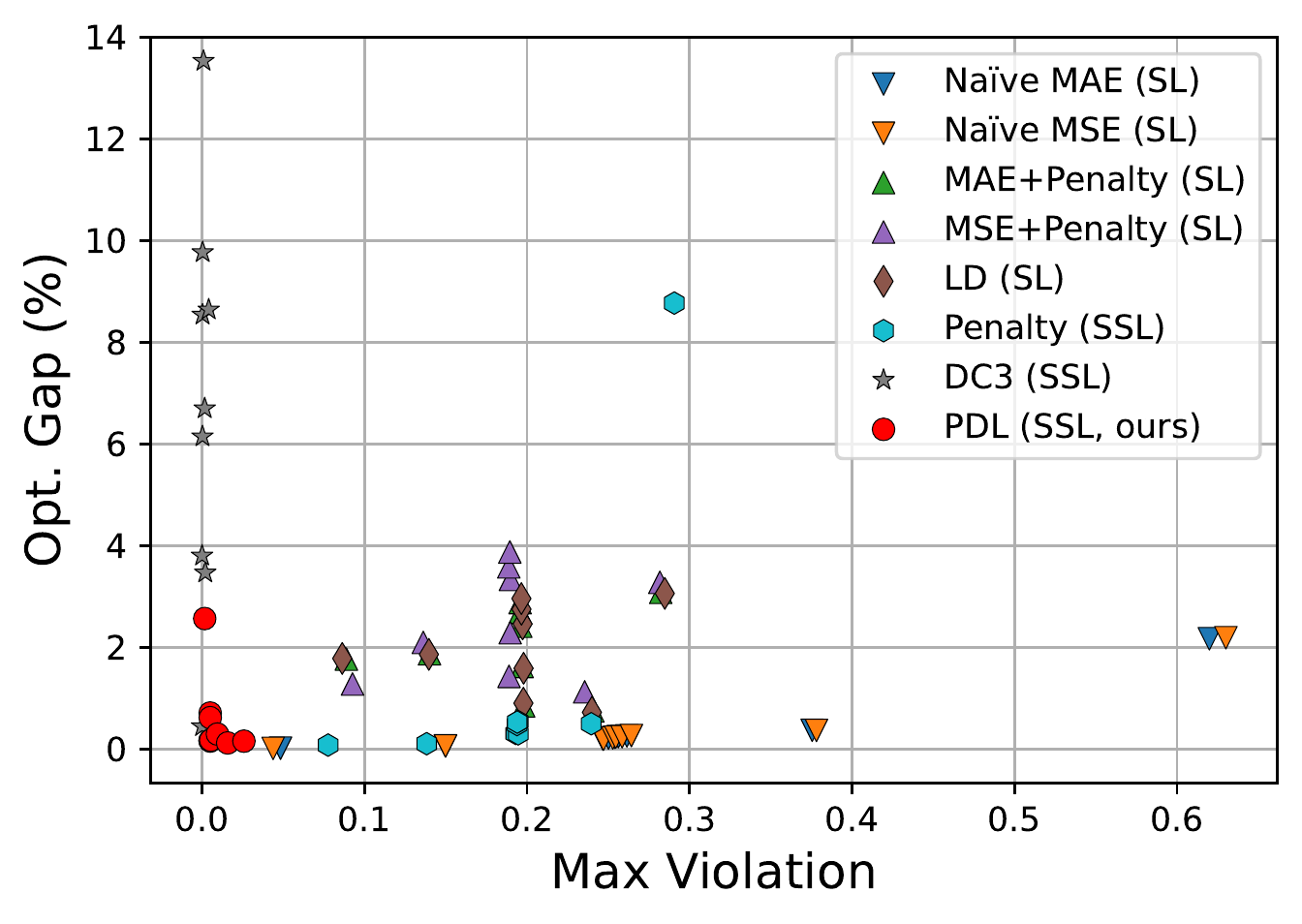}
\caption{The optimality gaps (\%) and maximum violations for various convex QP problems. 
Tested on 9 problem configurations using 
$n_{\text{eq}}\!=\!\{10,30,50,70,90\}$ and $n_{\text{ineq}}\!=\!50$, or $n_{\text{eq}}\!=\!50$ and $n_{\text{ineq}}\!=\!\{10,30,70,90\}$. 
The detailed results are described in Appendix~\ref{appx:add_exp_results}.
}
\label{fig:diverse_simple}
\end{figure}

Table~\ref{tab:result_qp} reports the performance results of the
convex QP problem and its non-convex variant on the test instances.
The first row for both tables reports the results of the ground truth,
which is obtained by the optimization systems OSQP
\cite{stellato2020osqp} or IPOPT \cite{wachter2006implementation}.  In
addition, the results of the ALM are also reported: they show that the
ALM converges well for these cases.  The optimality gap in percentage
is the average value of the optimality gaps over the test instances,
i.e., $\frac{|f(\ygt)-f(\by)|}{|f(\ygt)|}$.  In the tables, columns
`Max eq.', `Max ineq.', `Mean eq.', `Mean ineq.' represent the average
violation of the maximum or average values for the equality or
inequality constraints across the test instances.  The performance of
PDL, which is only slightly worse than ALM, is compelling when
considering both optimality and feasibility. The baselines each present
their own weakness.  For example, supervised approaches tend to
violate some constraints substantially, which can be marginally
mitigated by adding penalty terms or employing the LD method.  DC3, as
it exploits the completion and correction steps, satisfies the
constraints but presents significant optimality gaps.  The superiority
of PDL for these cases is even more compelling when testing with
various configurations.  Fig.~\ref{fig:diverse_simple} shows the
optimality gaps and the maximum violations in various settings.  Note
that the PDL results are located close to the origin, meaning that it
has negligible optimality gaps and constraint violations.

\subsubsection{Quadratically Constrained Quadratic Program (QCQP)}
\begin{table}[!t]
\centering
\small
\addtolength\tabcolsep{-3pt}
\begin{tabular}{@{}lcccc@{}}
\toprule
\multirow{2}{*}{Method} & \multirow{2}{*}{Obj.} & \multicolumn{2}{c}{Opt. Gap (\%)} & \multirow{2}{*}{Time (s)} \\
                        &                       & mean        & max                 &                           \\ \midrule
Opt.(Gurobi)            & 710.071               & -           &    -                &  257.361                  \\
Opt.(CVXPY-QCQP)        & 1082.793              & 52.597      & 408.631             &  4.646                    \\
Opt.(ALM)               & 710.923               & 0.119       & 7.659               &  128.992                  \\ \midrule
PDL(SSL, ours)          & \bb{711.064}          & \bb{0.141}  & \bb{8.307}          &  0.004                    \\ 
Penalty (SSL)           & 1295.388              & 82.555      & 109.406             &  0.004                    \\
\Naive{} MAE (SL)       & 716.132               & 0.834       & 67.051              &  0.004                    \\
\Naive{} MSE (SL)       & 716.695               & 0.911       & 85.073              &  0.004                    \\
MAE+Penalty (SL)        & 714.891               & 0.664       & 61.093              &  0.004                    \\
MSE+Penalty (SL)        & 717.869               & 1.072       & 80.577              &  0.004                    \\
LD (SL)                 & 715.522               & 0.751       & 73.290              &  0.004                    \\ \bottomrule
\end{tabular}
\caption{Performance results for the QCQP problem (Eq.~\eqref{eq:qcqp}) with $n=50$ and $n_{\text{aff}}=75$ on 1000 test instances. The best results are shown in bold.}
\label{tab:qcqp_result}
\end{table}

QCQP is basically a non-convex problem in which both objective function and
constraints are quadratic functions. This experiment considers a
binary least-square minimization problem, which can be defined as
\begin{equation}
\label{eq:qcqp}
    \min_{\by\in\RR^n} \left(\mathbf{A}\by\!-\!\bx\right)^T\!\!\left(\mathbf{A}\by\!-\!\bx\right),\text{ s.t. }y_i^2=1,\;\forall i\!\in\!\{1,\dots,n\},
\end{equation}
where $\mathbf{A}\in\RR^{n_{\text{aff}}\times n}$ and
$\bx\in\RR^{n_{\text{aff}}}$.  Here, $n_{\text{aff}}$ is the dimension
of the affine space.  This problem can be formulated as a 0-1 integer
programming problem as
\begin{equation*}
\min_{\by\in\{-1,1\}^n} \left(\mathbf{A}\by-\bx\right)^T\left(\mathbf{A}\by-\bx\right).
\end{equation*}

\noindent
Gurobi 9.5.2 \cite{gurobi} was used as the optimization solver to
produce the instance data for the supervised baselines, and also
served as the reference for computing optimality gaps.  In addition,
CVXPY-QCQP \cite{park2017general} was used as an alternative heuristic
optimizer for additional comparisons.
Since it is a heuristic method, CVXPY-QCQP tends to solve problems
faster, but often converges to a suboptimal solution.  The ALM is also
included in the table for comparison. As it is a local search method
and the local solution depends on the initial point, the table reports
the best objective values obtained from 50 independent ALM runs with
randomized initial points.
        
The overall results for $n=50$ and $n_{\text{aff}}=75$ are illustrated
in Table~\ref{tab:qcqp_result}.  The supervised baselines are trained
on 10,000 instances solved by Gurobi and the training instances for
the self-supervised learning methods are generated on the fly during
training.  In this experiment, finding a feasible point is readily
achievable as one can exploit a sigmoid function in the last layer of
the neural network.  Hence, the constraint violation results are not
reported. Similar to the ALM and also inspired by the stochastic
multi-start local search \cite{schoen1991stochastic}, 10 independent
models with different seeds were built for each training method, and
the reported results in the table are based on the best values among
10 inferences.  The times reported in the table denote the inference
times for training the learning schemes or the solving times for the
optimization runs.

The results in Table~\ref{tab:qcqp_result} show that PDL is
particularly effective. PDL is almost as accurate as the ALM in terms
of optimality gaps, with a mean optimality gap of 0.14\% and a maximum
optimality gap of only 8.31\% among 1,000 test instances. Note that
the ALM method has a mean optimality gap of 0.12\% and a maximum
optimality gap of only 7.66\% when solving (not learning) the 1,000
test instances. As a result, {\em PDL provides about an order of
  magnitude improvement over the baselines.}  Moreover, PDL presents a
speedup of around $64,340\times$ over Gurobi at the cost of a
reasonable loss in optimality.

\begin{table}[!t]
\centering
\small
\addtolength\tabcolsep{-4pt}
\begin{tabular}{@{}lcccc@{}}
\toprule
\multirow{2}{*}{Method} & \multicolumn{2}{c}{case57} & \multicolumn{2}{c}{case118} \\ \cmidrule(lr){2-3}\cmidrule(lr){4-5}
                        & Gap(\%) & Max viol.        & Gap(\%) & Max viol. \\ \midrule
PDL (SSL, ours)         & 0.21(0.02) & \bb{0.01}(0.00) & 0.73(0.12) & \bb{0.04}(0.01)\\ 
Penalty (SSL)           & 9.73(0.03) & 0.03(0.00) & 5.51(0.11) & 0.05(0.00) \\ \midrule
\Naive{} MAE (SL)       & 0.39(0.54) & 0.09(0.06) & 0.73(0.25) & 0.35(0.13) \\
\Naive{} MSE (SL)       & 0.06(0.01) & 0.06(0.03) & \bb{0.34}(0.38) & 0.21(0.12) \\
MAE+Penalty (SL)        & 0.37(0.52) & 0.03(0.01) & 0.91(0.86) & 0.08(0.02) \\
MSE+Penalty (SL)        & \bb{0.05}(0.01) & 0.02(0.00) & 0.37(0.33) & 0.05(0.01)\\
LD (SL)                 & 0.33(0.45) & 0.02(0.00) & 0.70(0.60) & 0.06(0.00)\\
\bottomrule
\end{tabular}
\caption{Performance results for the case56 and case118 AC-OPF problems. 
Std. dev. in parenthesis is evaluated across 5 independent runs. 
The detailed results on various AC-OPF problems are provided in Appendix~\ref{appx:add_exp_results}. 
The best values are in bold.}
\label{tab:result_acopf}
\end{table}

\subsubsection{AC-Optimal Power Flow Problems}

AC-optimal power flow (AC-OPF) is a fundamental building block for
operating electrical power systems.  AC-OPF determines the most
economical setpoints for each generator, while satisfying the demand
and physical/engineering constraints simultaneously. Different
instances of an AC-OPF problem are given by the active and reactive
power demands. The variables, which correspond to the output of the
(primal) neural network, are the active and reactive power generations,
the nodal voltage magnitudes, and the phase angles for all buses.  The
mathematical formulation of AC-OPF used in the paper is based on the
formulation from PGLIB \cite{babaeinejadsarookolaee2019power} and is
described in Appendix~\ref{appx:acopf}.

Table~\ref{tab:result_acopf} reports the performance results of the
case56 and case118 in PGLib
\cite{babaeinejadsarookolaee2019power}. The results show that PDL
produces competitive optimality gaps and maximum constraint
violations, while supervised penalty approaches also show fine
performances. However, for bulk power systems, where gathering the
pre-solved instances is cumbersome, PDL may become a strong
alternative to the supervised approaches for yielding real-time
approximations of the solutions.

\subsection{Training Time vs. Data Generation Time}

\begin{table}[!t]
\centering
\small
\addtolength\tabcolsep{-4pt}
\begin{tabular}{@{}lccc@{}}        
        \toprule
\multirow{2}{*}{Test Case}   & \multicolumn{2}{c}{Training Time (s)} & \multirow{2}{*}{\begin{tabular}{c}Pre-solved Training Set\\Generation Time (s)\end{tabular}} \\ \cmidrule(lr){2-3}
                             & PDL (SSL)      & LD (SL)                &                                               \\ \midrule
Convex QP                    & 1558.5   & 1493.3            & 32.5                                          \\
Nonconvex QP                 & 1624.5   & 1550.4            & 630.1                                         \\
AC-OPF(case57)               & 5932.5   & 8473.2            & 3690.2                                        \\
AC-OPF(case118)              & 7605.1   & 9149.6            & 13120.3                                       \\
QCQP                         & 5553.2   & 3572.3            & 2573607.3 $\approx$ 715 hours \\
\bottomrule
\end{tabular}
\caption{The averaged elapsed time (s) for training PDL and the representative supervised learning baseline, LD (mid-column). The CPU time for gathering the pre-solved training instances (right-column).}
\label{tab:training_time_main}
\end{table}

Table \ref{tab:training_time_main} reports the training times for PDL, 
and shows that they are comparable to those of one of the supervised
learning baselines, LD. 
Importantly, it also highlights a key benefit of PDL and
self-supervised methods in general. Supervised learning schemes rely
on pre-solved instances and may need to run an optimization solver on
each input data.  The QCQP problem required around $715$ hours of CPU
time to collect the training dataset.  In contrast, PDL, being
self-supervised, does not need this step.  This potential benefit may
be important for training input/output mappings for large constrained
optimization problems in a timely manner.

\section{Conclusion}
\label{sec:conclusion}

This paper presented PDL, a self-supervised Primal-Dual Learning
scheme to learn the input/output mappings of constrained optimization
problems. Contrary to supervised learning methods, PDL does not
require a collection of pre-solved instances or an optimization
solver. Instead, PDL tracks the iteration points of the ALM and 
jointly learns primal and dual networks that
approximate primal and dual solution estimates.
Moreover, being a primal-dual method, PDL features a loss function for
primal training with instance-specific multipliers for its constraint
terms. Experimental results on a number of constrained optimization
problems highlight that PDL finds solutions with small optimality gaps
and negligible constraint violations. Its approximations are
remarkably close to the optimal solutions from the ALM, and PDL
typically exhibits improved accuracy compared to supervised methods
and primal self-supervised approaches.

PDL has also a number of additional benefits that remain to be
explored.  Being a primal-dual method, PDL provides dual
approximations which can be useful for downstream tasks (e.g.,
approximating Locational Marginal Prices in power systems or prices in
general).  Dual solutions estimates can also be leveraged when
learning is used for warm-starting an optimization or, more generally,
speeding up an optimization solver by identifying the set of binding
constraints.
Finally, observe that
PDL is generic with respect to the type of underlying primal and dual neural
networks. The use of implicit layer to restore the feasibility of the primal solution estimates 
can be considered.
Exploring more complex architectures for a variety of graph
problems is also an interesting direction.

\section*{Acknowledgments}
This research is partly supported by NSF under Award Number 2007095 and 2112533, and ARPA-E, U.S. Department of Energy under Award Number DE-AR0001280.

\bibliography{refs}

\clearpage
\appendix
\section{Augmented Lagrangian Method}\label{appx:alm}

    ALM \cite{bertsekas2014constrained,hestenes1969multiplier,rockafellar1974augmented,andreani2008augmented,wright1999numerical} 
    has been used for solving nonlinear constrained optimization problems.
    Suppose that we aim at solving the following constrained optimization problem:
    \begin{equation*}
    \begin{aligned}
        &P(\by):&&\minimize_{\by\in\bY}       && f(\by),\\
	    &&&\text{\textbf{subject to }} && g_{j}(\by) \leq 0,\; \forall j\in\cG,\\
	    &&&                            && h_{j}(\by) = 0,\; \forall j\in\cH.
    \end{aligned}
    \end{equation*}
    Given the Lagrangian multiplier estimates $\lamg\in\RR^{|\cG|}$, $\lamh\in\RR^{|\cH|}$, 
    the augmented Lagrangian function $L_{\rho}\!:\!\RR^n\!\times\!\RR_+^{|\cG|}\!\times\!\RR^{|\cH|}\rightarrow \RR$ is defined as follows:
    \begin{equation*}
    \begin{aligned}
        L_\rho(\by,\lamg,\lamh) = f(\by) + \lamg^T\mathbf{g}(\by) + \lamh^T\mathbf{h}(\by) \\
        + \frac{\rho}{2}\left(\ltwonorm{\max\{\mathbf{g}(\by),0\}}^2 + \ltwonorm{\mathbf{h}(\by)}^2 \right),
    \end{aligned}
    \end{equation*}
    where the max operator is applied element-wisely. 
    At each iteration $k$, given the fixed Lagrangian multiplier estimates $\lamg_k$, $\lamh_k$, 
    it is aimed to solve the primal subproblem as
    \begin{equation*}
        \by_k \in \argmin L_\rho(\by, \lamg_k, \lamh_k)
    \end{equation*}
    Then, using the proximal algorithm the Lagrangian multiplier estimates are updated as 
    \begin{equation*}
    \begin{aligned}
        &\lamg_{k+1} \leftarrow \max\{\lamg_k+\rho\mathbf{g}(\by),0\}\\
        &\lamh_{k+1} \leftarrow \lamh_k+\rho\mathbf{h}(\by).
    \end{aligned}
    \end{equation*}
    Also, the scaler $\rho>0$ could be increased when the violation gets worse.
    The $\rho$ updating rule comes from \cite{andreani2008augmented}, 
    which considers both feasibility and complementarity.
    Specifically, the violation is first calculated for both equalities and inequalities as
    \begin{align}
        v_k = \max\{ \linfnorm{\mathbf{h}_\bx(\by)},\linfnorm{\bm{\sigma_\bx(\by)}} \}\label{eq:rho_violations_alm} \\
        \text{where }\sigma_{\bx,j}(\by)=\max\{g_{\bx,j}(\by), -\frac{\lambda_{k,j}}{\rho}\} ,\;\forall j\in\cG \nonumber 
    \end{align}
    Then, at every iteration $k>1$, the penalty coefficient $\rho$ is increased 
    when the maximum violation $v_k$ is greater than a tolerance value $\tau v_{k-1}$ as 
    \begin{equation}
    \label{eq:update_rho}
        \rho \leftarrow \min\{\alpha\rho,\rhomax\}\;\text{if } v_k > \tau v_{k-1}
    \end{equation}
    
    Overall, Algorithm~\ref{alg:alm} summarizes the procedure of the ALM.
    It is known that the primal and dual variables through the ALM converge
    to the KKT point at the limit.
    To show its performance, the ALM (Algorithm~\ref{alg:alm}) is applied
    to the synthetic problems in the experiments including the QC problems
    and QCQP problem.
    In the implementation, the function \texttt{minimize} in Scipy 1.8.1 
    is used for solving the primal subproblem (corresponding to the 
    line~\ref{line:primal_subproblem} in Algorithm~\ref{alg:alm}). 
    Specifically, a nonlinear conjugate gradient algorithm by Polak and 
    Ribiere \cite{wright1999numerical} involved in the function 
    \texttt{minimize} is utilized with the terminating tolerance of $1\mathrm{e}{\textrm{-}4}$.
    
    For the QC problems, $\alpha$ is set to $10$, initial penalty coefficient $\rho$ is set to $1.$, 
    and $\tau=0.5$. Also the maximum outer iteration $K$ is set to $20$.
    The Lagrangian Multiplier estimates are initialized to zero.
    The initial primal point is randomly sampled from the uniform 
    distribution over $[-1,1]$.
    
    For the QCQP problem, since it is highly non-convex, stochastic 
    multi-start local search \cite{schoen1991stochastic} is applied. 
    Specifically, 50 independent runs are conducted and the best objective value
    is chosen.
    To do so, the initial primal point is also i.i.d. sampled from the uniform
    distribution over $[-2,2]$. Also the initial Lagrangian multiplier estimates
    $\lamh$ are i.i.d sampled from the uniform distribution over $[-2,2]$. 
    (Note that $\lamg$ is not used in this problem.)
    The parameters are set differently; $\alpha=1.2$, $\rho=0.1$, $\tau=0.5$.
    The maximum outer iteration $K$ is set to $50$.
    
\begin{algorithm}[tb]
\caption{Augmented Lagrangian Method (ALM)}\label{alg:alm}
\begin{algorithmic}[1] 
\State Initialize $\lamg_1$, $\lamh_1$, and $\rho$
\For{$k \in \{1,\dots,K\}$}
    \State Solve $\by_k \in \argmin_\by L_\rho(\by, \lamg_{k}, \lamh_{k})$\label{line:primal_subproblem}
    \State $\displaystyle \lamg_{k+1} \leftarrow \max\{\lamg_k+\rho\mathbf{g}(\by),0\} $
    \State $\lamh_{k+1} \leftarrow \lamh_k+\rho\mathbf{h}(\by) $
    \State Calculate $v_k$ using Eq.~\eqref{eq:rho_violations_alm}
        \If{$v_k<\epsilon$} \State Converged
    \EndIf
    \State Update $\rho$ using Eq.~\eqref{eq:update_rho}
\EndFor
\end{algorithmic}
\end{algorithm}
 
\section{Details on AC-OPF Problem}
\label{appx:acopf}
    The \emph{Optimal Power Flow} (OPF) determines the most economic 
    generator setpoints that satisfy the load demands as well as 
    physical constraints simultaneously in a power network.
    The power network can be represented as a graph $(\cN,\cE)$ where 
    $\cN$ is the nodes representing the set of bus IDs, which also contains generators
    and load units, and $\cE$ represents the set of transmission line IDs between buses. 
    $(lij)\in\cE$ where $l$ is a branch ID connected from node $i$ to $j$.
    The set $\cE^R$ captures the reversed orientation of $\cE$, i.e., $(lji)\in\cE^R$.
    Also, two subsets of $\cN$ are defined; the set of generator IDs $\cG$ and set of load demand IDs $\cL$.
    
    Overall, the AC-OPF formulation is detailed in Model~\eqref{model:acopf_formulation}.
    This AC-OPF problem is a nonlinear and non-convex program over complex numbers and variables.
    AC power generation $S^g$ is a complex number of which a real part $p^g$ is 
    an active power generation and an imaginary part $q^g$ is a reactive power generation
    per each generator. Also, at each bus, AC voltage $V$ can be separated into a voltage
    magnitude $v$ and voltage angle $\theta$.
    
    The objective~\eqref{eq:acopf_obj} is to minimize the sum of the quadratic costs of the active power generations.
    At the reference bus $r$, the voltage angle is set to zero as defined in constraint~\eqref{eq:acopf_cnst_refva}.
    Maximum and minimum active and reactive power generation at each generation unit are captured in constraints~\eqref{eq:acopf_cnst_pgbound}, \eqref{eq:acopf_cnst_qgbound}.
    Similarly, the bounds of the voltage magnitude are set in constraint~\eqref{eq:acopf_cnst_vm}.
    Constraint~\eqref{eq:acopf_cnst_flow1}, \eqref{eq:acopf_cnst_flow2} represent the power flow at each branch that is governed by Ohm's law.
    Constraint~\eqref{eq:acopf_cnst_balance} ensures that at each bus the power balance should be maintained.
    Constraints~\eqref{eq:acopf_cnst_thermal_limit}, \eqref{eq:acopf_cnst_angle_limit} capture the branch thermal limit and branch voltage angle difference limit, respectively.
    
    \begin{model}[!t]
    {\footnotesize
    \caption{AC-Optimal Power Flow (AC-OPF) Problem}
    \label{model:acopf_formulation}
    \begin{flalign} 
    &\text{\textbf{Sets:}}&\nonumber\\
    &\text{bus IDs }\cN&\nonumber\\
    &\text{generator IDs }\cG&\nonumber\\
    &\text{load demand IDs }\cL&\nonumber\\
    &\text{transmission line IDs }\cE&\nonumber\\
    \noalign{\vskip5pt}
    &\text{\textbf{Parameters:}}\nonumber\\
    &\text{AC power demand } S_i^d=p_i^d+jq_i^d &\forall i\in\cL\nonumber\\
    &\text{cost coefficients } c_{2i},c_{1i},c_{0i} & \forall i\in\cG\nonumber\\
    &\text{reference bus ID } r\in\cN\nonumber\\
    &\text{generation bounds }\pgimin,\pgimax,\qgimin,\qgimax & \forall i\in\cG\nonumber\\
    &\text{series admittance } Y_l &\forall (lij)\in\cE\nonumber\\
    &\text{line charging susceptance } b_l^c &\forall (lij)\in\cE\nonumber\\
    &\text{transformer parameter } T_l &\forall (lij)\in\cE\nonumber\\
    &\text{bus shunt admittance } Y_i^s &\forall i\in\cN\nonumber\\
    &\text{thermal bound } \bar{s}_l &\forall (lij)\in\cE\nonumber\\
    &\text{angle difference bounds } \underline{\Delta\theta}_l,\overline{\Delta\theta}_l & \forall (lij)\in\cE\nonumber\\
    \noalign{\vskip5pt}
    &\text{\textbf{Variables:}}\nonumber\\
    &\text{AC power generation }S^g_i=\pgi+j\qgi &\forall i\in\cG&& \nonumber\\
    &\text{AC voltage }V_i=\vmi \angle \vai        &\forall i\in\cN&& \nonumber\\
    &\text{AC branch power flow }S_{lij}           &\forall (lij)\!\in\!\cE\!\cup\!\cE^R&& \nonumber\\
    \noalign{\vskip5pt}
    &\text{\textbf{Objective Function: }}\nonumber\\
    &\min \sum_{i\in\cG}{c_{2i}(\pgi)^2+c_{1i}\pgi+c_{0i}},\label{eq:acopf_obj}\\
    \noalign{\vskip5pt}
    &\text{\textbf{Constraints:}}\nonumber\\
    &\theta_r= 0,\label{eq:acopf_cnst_refva}\\
    \setcounter{equation}{17}
    &\pgimin\leq\pgi\leq\pgimax,\; &\forall i\in\cG\label{eq:acopf_cnst_pgbound} \tag{\theequation a}\\
    &\qgimin\leq\qgi\leq\qgimax,\; &\forall i\in\cG\label{eq:acopf_cnst_qgbound} \tag{\theequation b}\\
    &\underline{v}_i\leq v_i\leq\overline{v}_i, &\forall i\in\cN\label{eq:acopf_cnst_vm}\\
    \setcounter{equation}{19}
    &S_{lij}\!=\!\left(Y_l^*\!-\!j\frac{b_l^c}{2}\right)\frac{v_i^2}{|T_l|^2}-Y_l^*\frac{V_iV_j^*}{T_l}              &\forall (lij)\!\in\!\cE\label{eq:acopf_cnst_flow1}\tag{\theequation a}\\
    &S_{lji}\!=\!\left(Y_l^*\!-\!j\frac{b_l^c}{2}\right)v_i^2-Y_l^*\frac{V_i^*V_j}{T_l^*}              &\forall (lji)\!\in\!\cE^R\label{eq:acopf_cnst_flow2}\tag{\theequation b}\\
    &\sum_{k\in\cG_i}S_k^g\!\!-\!\!S_i^d\!\!-\!\!Y_i^s|V_i|^2=\!\!\!\!\!\!\!\!\sum_{(lij)\in\cE\cup\cE^R}\!\!\!\!S_{lij} &\forall i\in\cN\label{eq:acopf_cnst_balance} \\
    &|S_{lij}|\leq \bar{s}_l &\forall(lij)\!\in\!\cE\!\cup\!\cE^R\label{eq:acopf_cnst_thermal_limit}\\
    &\underline{\Delta\theta}_l\leq \vai-\vaj\leq \overline{\Delta\theta}_l&\forall (lij)\!\in\!\cE\label{eq:acopf_cnst_angle_limit}
    \end{flalign}
    }
    \end{model}

\section{Details on Experiment Settings}
\label{appx:exp_settings}
    \subsection{Data Generation}
    \label{appx:data_gen}
        \subsubsection{QP cases}
            As the protocol of the experiments for the convex QP and its non-convex variant follows that of DC3 \cite{donti2021dc3}, the overall data generation process is exactly the same as that in the paper. 
            For the sake of completeness, the general process of generating the data instances is presented as follows.
            The quadratic matrix $\mathbf{Q}$ involved in the objective function (both Eq.~\eqref{eq:simple_qp} and Eq.~\eqref{eq:nonconvex_qp}) is assumed to be diagonal and positive semi-definite, which is ensured by setting the all diagonal entries are i.i.d. sampled from the uniform distribution on $[0.,1.]$.
            Also, the elements in matrices $\mathbf{A}$ and $\mathbf{G}$ are i.i.d. sampled from the uniform distribution on $[0.,1.]$.
            The matrix $\mathbf{h}$ is determined to be $\mathbf{h}=\sum_j{|\mathbf{G}\mathbf{A}^+|_{ij}}$ where $\mathbf{A}^+$ is the Moore-Penrose pseudo-inverse of $\mathbf{A}$, i.e. $\mathbf{A}\mathbf{A}^+=\mathbf{I}$.
            Then this ensures that the problem is always feasible with the variables satisfying $\mathbf{A}\by=\bx$ because
            \begin{equation}
                \mathbf{G}\by\leq\mathbf{h} \Rightarrow \mathbf{G}\mathbf{A}^+\bx\leq \sum_j{|\mathbf{G}\mathbf{A}^+|_{ij}}
            \end{equation}
            which holds for every $\bx$ with $x_j\in[-1,1]$ for all $j$.
            Thus, $\bx$ is i.i.d. sampled from the uniform distribution $\cU(-1,1)$ to satisfy the inequality constraints.
            
            For training the supervised baselines on the convex cases, OSQC 0.6.2 is utilized. Also, for the non-convex cases, IPOPT (cyipopt 1.1.0) is used.
            For both optimizers, the default settings are used when gathering the data instances.
        
        \subsubsection{QCQP case}
            The each element of input $\bx\in\RR^{n_{\text{aff}}}$ is i.i.d. sampled from the uniform distribution $\cU(-1,1)$.
            Also, Ensure that $\mathbf{A}\in\RR^{n_{\text{aff}}\times n}$ is full column rank in which all the elements are i.i.d. sampled from the standard normal distribution. 
            The self-supervised baseline and PDL use the training data instances generated on the fly, whereas the supervised baselines used $10000$ pre-solved instances of $\bx$ and $\by$ pairs for training.
            At inference time, 1000 instances that are unseen during training are used for reporting the performance.
            Gurobi 9.5.2 is used for generating the pre-solved instances for testing as well as for training the supervised baselines.
        
        \subsubsection{AC-OPF cases}
            \begin{table}[!t]
            \centering
            \resizebox{0.99\columnwidth}{!}{
            \begin{tabular}{@{}l|cccccccc@{}}
            \toprule
            Test case   & dim($\bx$) & dim($\by$) & $n_{\text{ineq}}$ & $n_{\text{eq}}$ & $|\cN|$ & $|\cG|$ & $|\cL|$ & $|\cE|$\\
            \midrule
            case57\_ieee   & 84  & 151 & 160  & 114 & 57  & 7  & 42  & 80 \\
            case118\_ieee  & 198 & 412 & 372  & 236 & 118 & 54 & 99  & 186\\
            \bottomrule
            \end{tabular}}
            \caption{Specifications of the AC-OPF test cases.}
            \label{tab:acopf_dimension}
            \end{table}
            
            As shown in Table~\ref{tab:acopf_dimension}, two test cases varying the number of buses, generators, and load units are considered.
            
            For the AC-OPF test cases, the load demands, $p^d$, $q^d$, for all load units $\cL$ are perturbed to augment the problem data instances.
            In PGLIB \cite{babaeinejadsarookolaee2019power}, the original (reference) load demands is specified.
            Let us denote the concatenated vector of the original active and reactive load demands as $\hat{\bx}$ then the perturbed input $\bx$ is realized by multiplying so-called \emph{load factor} $\mathbf{f}$ as
            \begin{equation}
                \bx = \mathbf{f}\odot \hat{\bx},
            \end{equation}
            where $\odot$ is the element-wise multiplication.
            The load factor $\mathbf{f}$ is sampled from the truncated multivariate Gaussian distribution following \cite{zamzam2020learning} as
            \begin{equation}
                \mathbf{f} \sim \mathcal{T}\mathcal{N} \left( \mathbf{1}, \Sigma, (1-\mu)\mathbf{1}, (1+\mu)\mathbf{1} \right),
            \end{equation}
            where the first argument is the mean, $\Sigma$ denotes the covariance matrix, and the other two are lower and upper limits for truncation.
            In our experiment, we set $\mu$ to $0.3$ and the covariance matrix $\Sigma$ to the correlated covariance with the correlation parameter of $0.8$.
            
            Totally 30000 instances are generated and separated into train/valid/test datasets with a ratio of 10:1:1.
            The optimizer is included in PowerModels.jl \cite{8442948}, which is IPOPT 3.14.4 in general with MUMPS 5.4.1 as a linear solver. We have used the default settings for experiments.

\subsection{Architecture Design and Training Procedure}
\label{appx:archi_design} 
    \subsubsection{QC cases}
    For both the primal and dual nets in PDL and for other baselines, MLP with 2 hidden layers of size 500 is used.
    Also, a ReLU activation is attached to each layer.
    Batch normalization and Dropout layers are excluded because it is found that adding those degrades the performance. But for DC3, those are not excluded as used in the original work.
    For the baselines, the number of maximum epochs is set to $10000$, which is equivalent to using the maximum outer iteration of $K=10$ with the $500$ inner epochs (for each primal or dual learning) for PDL.

    \subsubsection{QCQP case}
    The architectural design for PDL and the baselines is the same as it is for QC cases except that the Sigmoid activation is attached to the final layer to yield a binary decision, which is, in turn, scaled to $\left[-1,1\right]$.
    The number of iterations for the self-supervised penalty method is set to $100000$ and equivalently, for the supervised baselines, the number of epochs is set to $2000$.
    Also, for PDL, the number of outer iterations is set to $10$ and $5000$ inner iterations for both primal and dual learning are used.

    \subsubsection{AC-OPF cases}
    The input and output of the (primal) neural network are defined as below:
    \begin{description}[itemindent=0.in,leftmargin=!,labelwidth=\widthof{$\;\;-$ Output  }]
      \item[$\;\;-$ Input] AC power demands $p_i^d,q_i^d$, $\forall i\in\cN$
      \item[$\;\;-$ Output] AC power generations $p_i^g,q_i^g$, $\forall i\in\cG$\\ 
                            AC voltage magnitudes $v_i$, $\forall i\in\cN$\\
                            AC voltage angle difference $\vai-\vaj$, $\forall (lij)\in\cE $
    \end{description}
    The AC power demands are all concatenated to serve an input $\bx$.
    The hidden representation is embedded using 2 layered MLPs. Each layer has $1.2\times$dim($\by$) nodes. 
    Then the hidden representation is served as an input of the `individual' net for outputting the four individual estimates; active, reactive generations, and the voltage magnitude and angle difference.
    All four individual nets have one hidden layer with nodes of the same size as the output dimension for each output.
    Also, ReLU activations are followed to all fully-connected layers.
    Since the AC power generations and voltage magnitudes have bound constraints, the Hardsigmoid activation, which outputs a value in a range of $[0.,1.]$, is attached to the last layer. Then, using the bound limits, the convex combination of both bounds using the output of the individual model would be the eventual output of the neural net. 
    Consequently, the bound constraints Eq.~\eqref{eq:acopf_cnst_pgbound}, \eqref{eq:acopf_cnst_qgbound}, \eqref{eq:acopf_cnst_vm}, and \eqref{eq:acopf_cnst_angle_limit} in AC-OPF problem (Model~\eqref{model:acopf_formulation}) are always satisfied without having the explicit penalty functions.
    Once the outputs from the four individual neural nets are furnished, the power flows $S_{lij}$ can be directly calculated from Ohm's law (Eq.~\eqref{eq:acopf_cnst_flow1} and \eqref{eq:acopf_cnst_flow2} in AC-OPF problem (Model~\eqref{model:acopf_formulation})).
    
    For PDL, the number of outer iterations is set to $10$ and the inner epochs are set to $250$ for both learning subproblems.
    For the self-supervised and supervised baseline, the number of epochs is set to $5000$.

\subsection{Hyperparameters}
\label{appx:hyperparam}
    For all cases and methods, the number of instances in the minibatch $|\batch|$ is set to $200$.
    Also, the training process is performed using Adam optimizer \cite{kingma2014adam} with a learning rate of $1\mathrm{e}{\textrm{-}4}$.
    For PDL, the learning rate is decayed by $0.99$ when the validation loss is bigger than the best validation loss in the primal or dual learning procedure.
    For QCQP cases, validation loss is evaluated every 500 inner iterations to determine the adjustment of the learning rate whereas for the other problem cases, it is evaluated every epoch.
    Since this learning rate decay is not beneficial to the baselines empirically (it leads to premature convergence in most cases), it is not applied to those.
    
    \subsubsection{QP cases}
    For DC3, all the hyperparameters suggested by the original work~\cite{donti2021dc3} are used.
    The hyperparameters for PDL and other baselines are tuned using the convex case through the grid search and those resultant combinations are also applied to the non-convex QP variants.
    For all methods, here is the list of the candidate hyperparameters where the chosen hyperparameters are in bold.
    \begin{itemize}
        \item PDL
        \begin{itemize}
            \item $\tau$: $0.5$, $0.6$, $0.7$, $\mathbf{0.8}$, $0.9$
            \item $\rho$: $0.1$, $\mathbf{0.5}$, $1.$, $10.$
            \item $\rhomax$: $1000.$, $\mathbf{5000.}$, $10000.$, $50000.$ 
            \item $\alpha$: $1.$, $1.5$, $2.$, $5.$, $\mathbf{10.}$
        \end{itemize}
        \item MAE+Penalty, MSE+Penalty, and LD (SL)
        \begin{itemize}
            \item $\rho_g$: $0.1$, $0.5$, $1.$, $\mathbf{5.}$, $10.$, $50.$
            \item $\rho_h$: $0.1$, $0.5$, $1.$, $\mathbf{5.}$, $10.$, $50.$
            \item LD updating epochs: $\mathbf{50.}$, $100$, $200$, $300$, $500$, $1000$
            \item LD step size: $1\mathrm{e}{0}$, $1\mathrm{e}{\textrm{-}1}$, $1e{\textrm{-}2}$, $\mathbf{1\mathrm{e}{\textrm{-}3}}$, $1\mathrm{e}{\textrm{-}4}$
        \end{itemize}
        \item Penalty (SSL)
        \begin{itemize}
            \item $\rho_g$: $0.1$, $0.5$, $1.$, $\mathbf{5.}$, $10.$, $50.$, $100.$
            \item $\rho_h$: $0.1$, $0.5$, $1.$, $\mathbf{5.}$, $10.$, $50.$, $100.$
         \end{itemize}   
    \end{itemize}

\subsubsection{QCQP case}
    All experiment settings and the architectural design for all baselines and the PDL are the same as those for the QP problems.
    Hyperparameters are as follows:
    \begin{itemize}
        \item PDL 
        \begin{itemize}
            \item $\tau$: $0.5$, $0.6$, $0.7$, $\mathbf{0.8}$, $0.9$
            \item $\rho$: $0.1$, $0.5$, $\mathbf{1.}$, $10.$
            \item $\rhomax$: $1000.$, $5000.$, $\mathbf{10000.}$, $50000.$ 
            \item $\alpha$: $1.$, $\mathbf{1.5}$, $2.$, $5.$, $10.$
        \end{itemize}
        \item MAE+Penalty and LD (SL)
        \begin{itemize}
            \item $\rho_g$: $\mathbf{0.1}$, $0.5$, $1.$, $5.$, $10.$, $50.$
            \item $\rho_h$: $\mathbf{0.1}$, $0.5$, $1.$, $5.$, $10.$, $50.$
            \item LD updating epochs: $\mathbf{50}$, $100$, $200$, $300$, $500$, $1000$
            \item LD step size: $\mathbf{1\mathrm{e}{0}}$, $1\mathrm{e}{\textrm{-}1}$, $1e{\textrm{-}2}$, $1\mathrm{e}{\textrm{-}3}$, $1\mathrm{e}{\textrm{-}4}$
        \end{itemize}
        \item MSE+Penalty (SL)
        \begin{itemize}
            \item $\rho_g$: $0.1$, $0.5$, $\mathbf{1.}$, $5.$, $10.$, $50.$
            \item $\rho_h$: $0.1$, $0.5$, $\mathbf{1.}$, $5.$, $10.$, $50.$
        \end{itemize}
        \item Penalty (SSL)
        \begin{itemize}
            \item $\rho_g$: $0.1$, $0.5$, $1.$, $5.$, $10.$, $50.$, $\mathbf{100.}$
            \item $\rho_h$: $0.1$, $0.5$, $1.$, $5.$, $10.$, $50.$, $\mathbf{100.}$
         \end{itemize}   
    \end{itemize}

\subsubsection{AC-OPF cases}
    Hyperparameter tuning for AC-OPF cases were conducted on case118, then applied the same hyperparameter combination to the case57 problem.
    \begin{itemize}
        \item PDL
        \begin{itemize}
            \item $\tau$: $0.5$, $0.6$, $0.7$, $\mathbf{0.8}$, $0.9$
            \item $\rho$: $0.1$, $0.5$, $\mathbf{1.}$, $10.$
            \item $\rhomax$: $1000.$, $5000.$, $\mathbf{10000.}$, $50000.$ 
            \item $\alpha$: $1.$, $1.5$, $\mathbf{2.}$, $5.$, $10.$
        \end{itemize}
        \item MAE+Penalty, MSE+Penalty, and LD (SL)
        \begin{itemize}
            \item $\rho_g$: $0.1$, $0.5$, $\mathbf{1.}$, $5.$, $10.$, $50.$
            \item $\rho_h$: $0.1$, $0.5$, $\mathbf{1.}$, $5.$, $10.$, $50.$
            \item LD updating epochs: $\mathbf{50.}$, $100$, $200$, $300$, $500$, $1000$
            \item LD step size: $\mathbf{1\mathrm{e}{0}}$, $1\mathrm{e}{\textrm{-}1}$, $1e{\textrm{-}2}$, $1\mathrm{e}{\textrm{-}3}$, $1\mathrm{e}{\textrm{-}4}$
        \end{itemize}
        \item Penalty (SSL)
        \begin{itemize}
            \item $\rho_g$: $0.1$, $0.5$, $\mathbf{1.}$, $5.$, $10.$, $50.$, $100.$
            \item $\rho_h$: $0.1$, $0.5$, $\mathbf{1.}$, $5.$, $10.$, $50.$, $100.$
         \end{itemize}   
    \end{itemize}

\newpage
\onecolumn
\section{Additional Experiment Results}
\label{appx:add_exp_results}

\subsection{Training Time vs. Data Generation Time}

\begin{table}[h!]
\centering
\small
\addtolength\tabcolsep{-4pt}
\resizebox{0.95\columnwidth}{!}{
\begin{tabular}{@{}lcccccccc@{}}
\toprule
Test Case            & PDL (SSL) & Penalty (SSL) & DC3 (SSL) & \Naive{} MAE (SL) & \Naive{} MSE (SL) & MAE+Penalty (SL) & MSE+Penalty (SL)  & LD (SL) \\ \midrule
convex QP            & 1558.45   & 1516.40       & 8099.22   & 1338.56           & 1329.03           & 1406.61          & 1417.23           & 1493.29 \\
non-convex QP variant & 1624.46   & 1576.17       & 6133.50   & 1382.40           & 1372.23           & 1460.79          & 1474.86           & 1550.42 \\
QCQP                 & 555.32    & 559.99        &    -      & 321.22            & 322.44            & 344.73           &  344.94           & 357.23 \\
AC-OPF(case57)       & 5932.47   & 8173.71       &    -      & 7058.20           & 7038.17           & 8471.24          & 8528.35           & 8473.21 \\
AC-OPF(case118)      & 7605.10   & 9433.96       &    -      & 7628.11           & 7598.09           & 9147.71          & 9179.24           & 9149.57 \\
\bottomrule
\end{tabular}
}
\caption{Averaged training GPU time (s) of the approaches for the various test cases.}
\label{tab:training_time}
\end{table}

\begin{table}[h!]
\centering
\small
\begin{tabular}{@{}lccc@{}}
\toprule
Test Case                      & Instance Solving Time (s) & \# Training Instances & \multirow{2}{*}{\begin{tabular}{c}Pre-solved Training Set\\Generation Time (s)\end{tabular}} \\ 
                               &                           &                       & \\ \midrule
convex QP (OSQP)               & 0.0039                    & 8334                  & 32.50 \\
non-convex QP variant (IPOPT)   & 0.0756                    & 8334                  & 630.05 \\
AC-OPF (case57)                & 0.1476                    & 25000                 & 3690.21 \\
AC-OPF (case118)               & 0.5248                    & 25000                 & 13120.30 \\
QCQP (Gurobi)                  & 257.3607                  & 10000                 & 2573607.34 \\
\bottomrule
\end{tabular}
\caption{Averaged CPU time (s) for solving a problem instance. Considering the number of training instances, the CPU time for arranging the training dataset for the supervised learning schemes is calculated by default.}
\label{tab:data_generation_time}
\end{table}

Typically, DC3 takes a longer time to train because it involves the implicit differentiation (in the completion steps) to satisfy the equalities. The difference in time taken for training between PDL and Penalty (SSL) comes from the architectural design; PDL uses two independent neural networks and the number of hidden nodes varies depending on the dimension of the output (the dimension of the primal or dual solution space). When comparing the \naive{} SL methods with the penalty SL methods, it is obvious that the penalty ones take a longer time. This is because it involves the back-propagation through the constraint functions.

\subsection{Additional QP Results}

\begin{figure}[h!]
\centering
\includegraphics[width=.6\columnwidth]{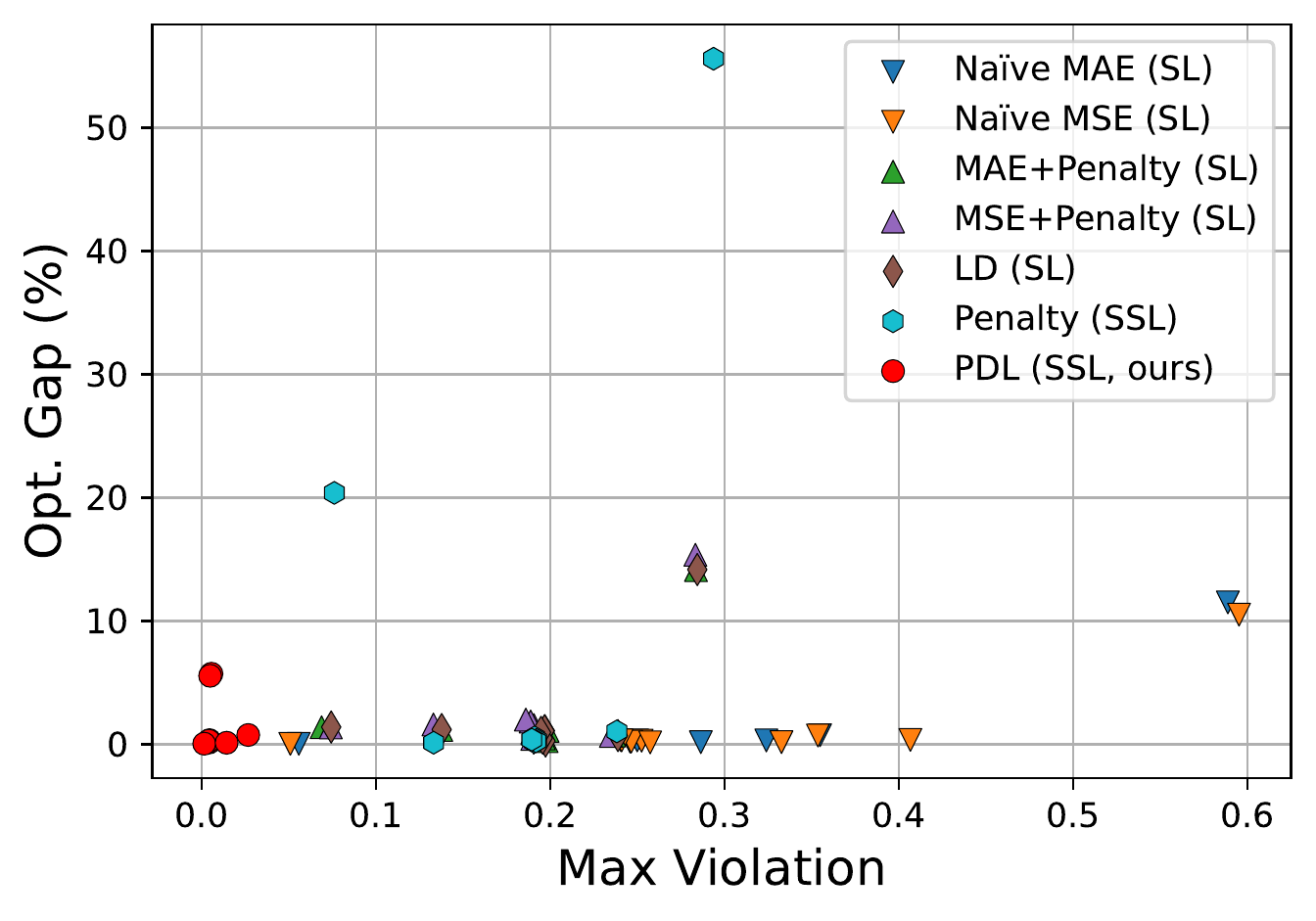}
\caption{The optimality gaps (\%) and maximum violations for various non-convex QP variants. The maximum violation means the maximum value between the equality and inequality constraint maximum violations. Tested on various problems generated using 
$n_{\text{eq}}\!=\!\{10,30,50,70,90\}$ and $n_{\text{ineq}}\!=\!50$, or $n_{\text{eq}}\!=\!50$ and $n_{\text{ineq}}\!=\!\{10,30,70,90\}$. DC3 is excluded due to its poor performance.}
\label{fig:diverse_nonconvex}
\end{figure}

\begin{table*}[ht!]
\centering
\resizebox{0.95\textwidth}{!}{
\begin{tabular}{@{}ll|cccccccc|cccccccc@{}}
\toprule
Method & $n_{\text{eq}}$    & $10$ & $30$ & $70$ & $90$ & $50$ & $50$ & $50$ & $50$ & $10$ & $30$ & $70$ & $90$ & $50$ & $50$ & $50$ & $50$ \\ 
       & $n_{\text{ineq}}$  & $50$ & $50$ & $50$ & $50$ & $10$ & $30$ & $70$ & $90$& $50$ & $50$ & $50$ & $50$ & $10$ & $30$ & $70$ & $90$ \\ \midrule
\multicolumn{2}{l|}{ } & \multicolumn{8}{c|}{convex QP problems} & \multicolumn{8}{c}{non-convex QP problems}\\
\multirow{1}{*}{Optimizer(OSQP or IPOPT)}        & Obj.         & -27.26 & -23.13 & -14.80 & -4.79 & -17.34 & -16.33 & -14.61 & -14.26& -18.57 & -16.92 & -7.13 & -1.33 & -18.11 & -12.30 & -13.32 & -8.64 \\\midrule
\multirow{4}{*}{PDL (SSL, ours)}  & Obj.         & -26.55 & -23.06 & -14.78 & -4.80 & -17.31 & -16.30 & -14.50 & -14.17& -18.56 & -15.96 & -7.12 & -1.33 & -18.07 & -11.57 & -13.29 & -8.61 \\
                                  & Opt. Gap(\%) & 2.57 & 0.3 & 0.12 & 0.16 & 0.16 & 0.17 & 0.71 & 0.62&  0.04 & 5.54 & 0.11 & 0.74 & 0.13 & 5.71 & 0.19 & 0.28 \\
                                  & Max eq.      & 0.00 & 0.01 & 0.02 & 0.03 & 0.01 & 0.00 & 0.00 & 0.01& 0.00 & 0.00 & 0.01 & 0.03 & 0.00 & 0.01 & 0.00 & 0.00 \\
                                  & Max ineq.    & 0.00 & 0.00 & 0.00 & 0.00 & 0.00 & 0.00 & 0.00 & 0.00& 0.00 & 0.00 & 0.00 & 0.00 & 0.00 & 0.00 & 0.00 & 0.00\\\midrule
\multirow{4}{*}{Penalty (SSL)}    & Obj.         & -27.28 & -23.11 & -14.76 & -5.30 & -17.39 & -16.32 & -14.52 & -14.18& -14.78 & -16.91 & -7.23 & -2.04 & -18.09 & -12.19 & -13.32 & -8.65 \\
                                  & Opt. Gap(\%) & 0.08 & 0.11 & 0.50 & 8.77 & 0.31 & 0.30 & 0.54 & 0.53& 20.38 & 0.12 & 1.03 & 55.60 & 0.18 & 0.54 & 0.23 & 0.37 \\
                                  & Max eq.      & 0.08 & 0.14 & 0.24 & 0.29 & 0.19 & 0.19 & 0.19 & 0.19& 0.08 & 0.13 & 0.24 & 0.29 & 0.19 & 0.19 & 0.19 & 0.19 \\
                                  & Max ineq.    & 0.05 & 0.04 & 0.00 & 0.00 & 0.02 & 0.02 & 0.02 & 0.02& 0.03 & 0.02 & 0.00 & 0.00 & 0.00 & 0.02 & 0.02 & 0.02 \\\midrule
\multirow{4}{*}{DC3 (SSL)}        & Obj.         & -26.30 & -21.11 & -14.23 & -4.76 & -16.14 & -14.90 & -12.60 & -12.85& -13.83 & -14.57 & 4e33 & -1.25 & -17.83 & -10.97 & -12.30 & -8.20 \\
                                  & Opt. Gap(\%) & 3.47 & 8.64 & 3.80 & 0.45 & 6.70 & 8.54 & 13.53 & 9.77& 25.53 & 13.55 & 4e34 & 3.67 & 1.48 & 10.45 & 7.49 & 5.00 \\
                                  & Max eq.      & 0.00 & 0.00 & 0.00 & 0.00 & 0.00 & 0.00 & 0.00 & 0.00& 0.00 & 0.00 & 2e2 & 0.00 & 0.00 & 0.00 & 0.00 & 0.00 \\
                                  & Max ineq.    & 0.00 & 0.00 & 0.00 & 0.00 & 0.00 & 0.00 & 0.00 & 0.00& 0.00 & 0.01 & 2e17 & 0.00 & 0.00 & 0.01 & 0.00 & 0.00 \\\midrule
\multirow{4}{*}{\Naive{} MAE (SL)} & Obj.        & -27.25 & -23.14 & -14.84 & -4.89 & -17.35 & -16.35 & -14.63 & -14.29& -18.57 & -16.93 & -7.18 & -1.48 & -18.13 & -12.32 & -13.35 & -8.66 \\
                                  & Opt. Gap(\%) & 0.02 & 0.06 & 0.37 & 2.17 & 0.20 & 0.22 & 0.24 & 0.26& 0.03 & 0.10 & 0.72 & 11.50 & 0.15 & 0.29 & 0.21 & 0.31 \\
                                  & Max eq.      & 0.05 & 0.15 & 0.38 & 0.62 & 0.25 & 0.25 & 0.26 & 0.26& 0.06 & 0.19 & 0.35 & 0.59 & 0.29 & 0.32 & 0.25 & 0.24 \\
                                  & Max ineq.    & 0.01 & 0.01 & 0.03 & 0.00 & 0.01 & 0.03 & 0.05 & 0.06& 0.01 & 0.05 & 0.01 & 0.00 & 0.02 & 0.07 & 0.04 & 0.04 \\\midrule
\multirow{4}{*}{\Naive{} MSE (SL)} & Obj.        & -27.25 & -23.14 & -14.84 & -4.89 & -17.35 & -16.35 & -14.63 & -14.29& -18.57 & -16.93 & -7.18 & -1.46 & -18.13 & -12.31 & -13.34 & -8.66 \\
                                  & Opt. Gap(\%) & 0.01 & 0.06 & 0.37 & 2.19 & 0.19 & 0.22 & 0.24 & 0.26& 0.02 & 0.13 & 0.69 & 10.50 & 0.15 & 0.33 & 0.20 & 0.30 \\
                                  & Max eq.      & 0.04 & 0.15 & 0.38 & 0.63 & 0.25 & 0.25 & 0.26 & 0.26& 0.05 & 0.26 & 0.35 & 0.60 & 0.33 & 0.41 & 0.25 & 0.24 \\
                                  & Max ineq.    & 0.01 & 0.01 & 0.03 & 0.00 & 0.01 & 0.02 & 0.05 & 0.06& 0.01 & 0.07 & 0.00 & 0.00 & 0.04 & 0.07 & 0.04 & 0.04 \\\midrule
\multirow{4}{*}{MAE+Penalty (SL)} & Obj.         & -26.77 & -22.69 & -14.66 & -4.99 & -17.16 & -16.04 & -14.18 & -13.82& -18.31 & -16.72 & -7.17 & -1.52 & -18.05 & -12.13 & -13.24 & -8.54 \\
                                  & Opt. Gap(\%) & 1.77 & 1.87 & 0.75 & 3.08 & 0.85 & 1.62 & 2.69 & 2.88& 1.37 & 1.16 & 0.70 & 14.12 & 0.24 & 1.07 & 0.47 & 0.97 \\
                                  & Max eq.      & 0.09 & 0.14 & 0.24 & 0.28 & 0.20 & 0.20 & 0.20 & 0.20& 0.07 & 0.14 & 0.24 & 0.28 & 0.20 & 0.20 & 0.20 & 0.19 \\
                                  & Max ineq.    & 0.00 & 0.00 & 0.00 & 0.00 & 0.00 & 0.00 & 0.00 & 0.00 & 0.00 & 0.00 & 0.00 & 0.00 & 0.00 & 0.00 & 0.00 & 0.00 \\\midrule
\multirow{4}{*}{MSE+Penalty (SL)} & Obj.         & -26.91 & -22.63 & -14.59 & -5.01 & -17.07 & -15.93 & -14.05 & -13.67& -18.31 & -16.64 & -7.14 & -1.53 & -18.01 & -12.08 & -13.17 & -8.44 \\
                                  & Opt. Gap(\%) & 1.28 & 2.10 & 1.13 & 3.28 & 1.43 & 2.29 & 3.58 & 3.88& 1.36 & 1.59 & 0.67 & 15.37 & 0.42 & 1.55 & 0.92 & 1.99 \\
                                  & Max eq.      & 0.09 & 0.14 & 0.24 & 0.28 & 0.19 & 0.19 & 0.19 & 0.19& 0.07 & 0.13 & 0.23 & 0.28 & 0.19 & 0.19 & 0.19 & 0.19 \\
                                  & Max ineq.    & 0.00 & 0.00 & 0.00 & 0.00 & 0.00 & 0.00 & 0.00 & 0.00 & 0.00 & 0.00 & 0.00 & 0.00 & 0.00 & 0.00 & 0.00 & 0.00 \\\midrule
\multirow{4}{*}{LD (SL)}          & Obj.         & -26.77 & -22.69 & -14.67 & -4.99 & -17.15 & -16.05 & -14.17 & -13.81& -18.31 & -16.71 & -7.17 & -1.51 & -18.05 & -12.13 & -13.24 & -8.54 \\
                                  & Opt. Gap(\%) & 1.78 & 1.86 & 0.72 & 3.06 & 0.90 & 1.59 & 2.75 & 2.96& 1.39 & 1.18 & 0.69 & 14.16 & 0.26 & 1.09 & 0.49 & 0.96 \\
                                  & Max eq.      & 0.09 & 0.14 & 0.24 & 0.28 & 0.20 & 0.20 & 0.20 & 0.20& 0.07 & 0.14 & 0.24 & 0.28 & 0.20 & 0.20 & 0.20 & 0.19 \\
                                  & Max ineq.    & 0.00 & 0.00 & 0.00 & 0.00 & 0.00 & 0.00 & 0.00 & 0.00 & 0.00 & 0.00 & 0.00 & 0.00 & 0.00 & 0.00 & 0.00 & 0.00 \\
\bottomrule
\end{tabular}
}
\caption{Performance results of various convex and non-convex QP problems.}
\label{tab:diverse_simple_nonconvex}
\end{table*}

We exclude DC3 results from Figure~\ref{fig:diverse_nonconvex} for the non-convex QP variant cases because it is unstable in some cases. The detailed performance values can be found in Table~\ref{tab:diverse_simple_nonconvex}.

\newpage
\subsection{Detailed QCQP results}
    
    \begin{table*}[h]
    \small
    \centering
    \resizebox{0.85\textwidth}{!}{
    \begin{tabular}{@{}l|cccccc@{}}
    \toprule
    \multirow{2}{*}{Method} & \multirow{2}{*}{Obj.}   & \multicolumn{4}{c}{Opt Gap (\%)}                                                                       & \multirow{2}{*}{Time (s)} \\
                            &                         & min                     & mean                    & median                    & max                    &\\ \midrule
    Optimizer(Gurobi)       & 710.071                 &  -                      & -                       & -                         & -                      & 257.361 \\
    Optimizer(CVXPY-QCQP)   & 1082.793                & 5.583                   & 52.597                  & 50.589                    & 408.631                & 4.646 \\
    Optimizer(ALM)          & 710.923                 & 0.000                   & 0.119                   & 0.000                     & 7.659                  & 128.992 \\ \midrule
    PDL(SSL, ours)          & \textbf{711.064}(0.049) & \textbf{0.000}(0.000)   & \textbf{0.141}(0.007)   & \textbf{0.000}(0.000)     & \textbf{8.307}(0.000)  & 0.004(0.000) \\ 
    Penalty (SSL)           & 1295.388(9.994)         & 62.211(2.179)           & 82.555(1.406)           & 82.145(1.418)             & 109.406(3.712)         & 0.004(0.000) \\
    \Naive{} MAE (SL)       & 716.132(0.544)          & \textbf{0.000}(0.000)   & 0.834(0.075)            & \textbf{0.000}(0.000)     & 67.051(14.249)         & 0.004(0.000) \\
    \Naive{} MSE (SL)       & 716.695(1.223)          & \textbf{0.000}(0.000)   & 0.911(0.168)            & \textbf{0.000}(0.000)     & 85.073(24.454)         & 0.004(0.000) \\
    MAE+Penalty (SL)        & 714.891(0.604)          & \textbf{0.000}(0.000)   & 0.664(0.082)            & \textbf{0.000}(0.000)     & 61.093(2.492)          & 0.004(0.000) \\
    MSE+Penalty (SL)        & 717.869(0.945)          & \textbf{0.000}(0.000)   & 1.072(0.130)            & \textbf{0.000}(0.000)     & 80.577(8.594)          & 0.004(0.000) \\
    LD (SL)                 & 715.522(0.696)          & \textbf{0.000}(0.000)   & 0.751(0.095)            & \textbf{0.000}(0.000)     & 73.290(10.249)         & 0.004(0.000) \\
    \bottomrule
    \end{tabular}
    }
    \caption{Detailed performance results for QCQP problem (Eq.~\eqref{eq:qcqp}) with $n=50$ and $n_{\text{aff}}=75$ on 1000 test instances. The best results are shown in bold. Std. dev. values in parenthesis calculated across 5 independent experiment sets. For the PDL and baselines, 10 independent neural networks are arranged and denote the best objective values per each experiment set.}
    \label{tab:qcqp_result_detail}
    \end{table*}
    
\subsection{Detailed AC-OPF results}
    
    \begin{table*}[!ht]
    \centering
    \small
    \begin{tabular}{@{}ll|cccc@{}}
    \toprule
    Method                            &              & case57\_ieee      & case118\_ieee      & case57\_ieee (SAD) & case118\_ieee (SAD) \\ \midrule
    \multirow{1}{*}{Optimizer(IPOPT)} & Obj.         & 36926.823         & 97422.703          & 36711.670          & 98451.014 \\ \midrule
    \multirow{4}{*}{PDL (SSL, ours)}  & Obj.         & 36891.156(16.770) & 97371.053(350.655) & 36567.166(107.837) & 93842.756(131.439)  \\
                                      & Opt. Gap(\%) & 0.206(0.015)      & 0.730(0.118)       & 0.538(0.115)       & 4.455(0.110)  \\
                                      & Max eq.      & 0.012(0.000)      & 0.041(0.005)       & 0.012(0.000)       & 0.045(0.008)  \\
                                      & Max ineq.    & 0.000(0.000)      & 0.001(0.000)       & 0.000(0.000)       & 0.000(0.000)  \\ \midrule
    \multirow{4}{*}{Penalty (SSL)}    & Obj.         & 33346.842(12.062) & 92020.227(106.900) & 32675.748(25.923)  & 88659.323(69.800) \\
                                      & Opt. Gap(\%) & 9.727(0.034)      & 5.510(0.110)       & 11.015(0.071)      & 9.677(0.072) \\
                                      & Max eq.      & 0.025(0.000)      & 0.047(0.004)       & 0.026(0.001)       & 0.045(0.003) \\
                                      & Max ineq.    & 0.000(0.000)      & 0.001(0.000)       & 0.000(0.000)       & 0.000(0.000) \\ \midrule
    \multirow{4}{*}{\Naive{}-MAE (SL)} & Obj.        & 36873.743(46.728) & 96882.388(375.567) & 36753.908(50.397)  & 98591.628(164.089)  \\
                                      & Opt. Gap(\%) & 0.387(0.538)      & 0.725(0.252)       & 1.052(1.915)       & 0.334(0.086)  \\
                                      & Max eq.      & 0.089(0.059)      & 0.349(0.125)       & 0.193(0.140)       & 0.324(0.039)  \\
                                      & Max ineq.    & 0.000(0.000)      & 0.009(0.007)       & 0.000(0.000)       & 0.004(0.003)  \\ \midrule
    \multirow{4}{*}{\Naive{}-MSE (SL)} & Obj.        & 36928.445(11.403) & 97105.196(395.409) & 36717.866(19.535)  & 98530.690(262.072)  \\
                                      & Opt. Gap(\%) & 0.056(0.014)      & 0.337(0.380)       & 0.071(0.023)       & 0.240(0.194)  \\
                                      & Max eq.      & 0.058(0.026)      & 0.206(0.118)       & 0.072(0.025)       & 0.264(0.032)  \\
                                      & Max ineq.    & 0.000(0.000)      & 0.007(0.004)       & 0.000(0.000)       & 0.004(0.002)  \\ \midrule
    \multirow{4}{*}{MAE+Penalty (SL)} & Obj.         & 36874.845(40.974) & 96620.466(971.152) & 36697.187(25.572)  & 98198.849(839.035) \\
                                      & Opt. Gap(\%) & 0.369(0.518)      & 0.914(0.858)       & 0.621(0.457)       & 0.696(0.762) \\
                                      & Max eq.      & 0.031(0.006)      & 0.076(0.018)       & 0.039(0.006)       & 0.099(0.011) \\
                                      & Max ineq.    & 0.000(0.000)      & 0.000(0.000)       & 0.000(0.000)       & 0.000(0.000) \\ \midrule
    \multirow{4}{*}{MSE+Penalty (SL)} & Obj.         & 36925.534(9.584)  & 97172.067(431.041) & 36720.188(24.651)  & 98547.417(229.001) \\
                                      & Opt. Gap(\%) & 0.048(0.008)      & 0.372(0.332)       & 0.461(0.461)       & 0.301(0.081) \\
                                      & Max eq.      & 0.017(0.003)      & 0.050(0.006)       & 0.018(0.001)       & 0.055(0.003) \\
                                      & Max ineq.    & 0.000(0.000)      & 0.000(0.000)       & 0.000(0.000)       & 0.000(0.000) \\ \midrule
    \multirow{4}{*}{LD (SL)}          & Obj.         & 36907.882(27.457) & 96763.095(655.200) & 36703.457(29.607)  & 98546.519(575.095) \\
                                      & Opt. Gap(\%) & 0.326(0.451)      & 0.698(0.596)       & 0.631(0.444)       & 0.590(0.307) \\
                                      & Max eq.      & 0.022(0.002)      & 0.059(0.004)       & 0.029(0.003)       & 0.067(0.004) \\
                                      & Max ineq.    & 0.000(0.000)      & 0.001(0.001)       & 0.000(0.000)       & 0.000(0.000) \\
    \bottomrule
    \end{tabular}
    \caption{Performance results for the AC-OPF test cases.}
    \label{tab:acopf_performance}
    \end{table*}
    
    Small Angle Difference (SAD) variants of the original AC-OPF cases are also included in the AC-OPF test cases. SAD variants are more challenging than the original ones because the branch voltage angle difference is tightened in SAD cases. As shown in Table~\ref{tab:acopf_performance}, the violations get worse in most SAD cases.

\end{document}

%% file: commands.tex
\usepackage{algorithm,algorithmicx,algpseudocode}
\usepackage{amsmath,amsfonts,bm}
\usepackage{todonotes}
\usepackage{enumitem}
\usepackage{calc}
\usepackage{booktabs}
\usepackage{multirow}
\usepackage{lipsum}  
\usepackage{romannum}
\usepackage{multibib}
\newcites{appx}{References (Appendix)}

\newcommand{\ub}[1]{\underbar{#1}}
\newcommand{\bb}[1]{\textbf{#1}}

\def\thmspace{0.2em}

\newcommand{\cE}{\mathcal{E}}
\newcommand{\cH}{\mathcal{H}}
\newcommand{\cG}{\mathcal{G}}
\newcommand{\cN}{\mathcal{N}}

\newcommand{\cL}{\mathcal{L}}
\newcommand{\cU}{\mathcal{U}}

\newcommand{\RR}{\mathbb{R}}

\DeclareMathOperator*{\argmin}{arg\!\min}
\newcommand{\minimize}{\operatornamewithlimits{\text{\textbf{minimize}}}}

\def\naive{na\"{i}ve}
\def\Naive{Na\"{i}ve}

\newcommand{\norm}[1]{\left\lVert#1\right\rVert} 
\newcommand{\ltwonorm}[1]{\left\lVert#1\right\rVert_2} 
\newcommand{\linfnorm}[1]{\left\lVert#1\right\rVert_\infty} 

\newcommand{\ygt}{\mathbf{y}^*} 

\newcommand{\bx}{\mathbf{x}} 
\newcommand{\by}{\mathbf{y}} 
\newcommand{\dataset}{\mathcal{D}}
\newcommand{\batch}{\mathcal{B}}

\newcommand{\pgi}{p_i^g}
\newcommand{\qgi}{q_i^g}
\newcommand{\pgimin}{\underline{p}_i^g}
\newcommand{\pgimax}{\overline{p}_i^g}
\newcommand{\qgimin}{\underline{q}_i^g}
\newcommand{\qgimax}{\overline{q}_i^g}
\newcommand{\vmi}{v_i}

\newcommand{\vai}{\theta_i}
\newcommand{\vaj}{\theta_j}

\newcommand{\bg}{\mathbf{g}} 
\newcommand{\bh}{\mathbf{h}} 
\newcommand{\bY}{\mathbf{Y}} 

\newcommand{\lamg}{\bm{\mu}} 
\newcommand{\lamh}{\bm{\lambda}}
\newcommand{\viol}[1]{\nu\left(#1\right)}
\newcommand{\rhomax}{\rho_{\text{max}}}

\floatstyle{ruled}
\newfloat{model}{thp}{lop}
\floatname{model}{Model}